\documentclass{article}
\usepackage[T1]{fontenc}
\usepackage{times}
\usepackage[margin=1in]{geometry}
\usepackage{natbib}

\usepackage{amsmath,amsfonts,bm}

\def\eqref#1{equation~\ref{#1}}

\def\1{\bm{1}}

\DeclareMathAlphabet{\mathsfit}{\encodingdefault}{\sfdefault}{m}{sl}
\SetMathAlphabet{\mathsfit}{bold}{\encodingdefault}{\sfdefault}{bx}{n}

\usepackage{booktabs}
\usepackage{hyperref}
\usepackage{url}
\usepackage{graphicx}
\usepackage{float}
\date{}

\title{Explaining Deep EEG Emotion Classifiers with Central Counterfactuals}

\author{\textbf{Vitor Inserra\textsuperscript{1},
Amit Chalmeti\textsuperscript{2}, Pranav Mucharla\textsuperscript{1}}, \\
\textbf{Yashasree Gadipalli\textsuperscript{1}, Neha Panda\textsuperscript{3}, Sai Nagamalla\textsuperscript{1}, Gabriel Matos\textsuperscript{1},} \\
\textbf{Rosie Dutt\textsuperscript{3},
Raghavendra Pradyumna Pothukuchi\textsuperscript{1},
Richard L. Marks\textsuperscript{1}} \\[0.5em]
\textsuperscript{1}Department of Computer Science, UNC Chapel Hill \\
\textsuperscript{2}Irving Medical Center, Columbia University \\
\textsuperscript{3}Department of Psychology and Neuroscience, UNC Chapel Hill \\[0.5em]
\texttt{inserra@cs.unc.edu}
}

\begin{document}

\maketitle
\begin{abstract}
Counterfactual explanations of electroencephalography (EEG) deep learning models may provide actionable insights for downstream applications. However, they can be incompatible with target-class distributions and may produce unrealistic EEG patterns. We propose a framework for generating counterfactual explanations under distributional and physiological constraints. A joint variational classifier-decoder combines affect prediction and EEG reconstruction with adversarial regularization to reduce subject-specific information. The counterfactual objective combines target-class prediction loss, class-conditional density regularization, proximity penalties in latent and reconstructed feature spaces, and physiological constraints on decoded EEG. Typicality optimization aims to generate counterfactual explanations central to the learned target-class distribution in latent space and express the corresponding changes in interpretable EEG frequency-band features. On DREAMER, we averaged ten window probabilities per trial. In leave-one-subject-out evaluation, EEG reconstruction achieved $R^2\simeq0.47$ and trial-level recognition of self-reported emotion was above chance. For eligible, correctly classified trials, we decoded counterfactuals and re-encoded the resulting EEG windows. Full counterfactual optimization (CFO) combined typicality regularization with additional refinement and achieved decoded target-class validity of 67.3\% for arousal and 88.8\% for valence, versus 24.4\% and 48.6\% for base CFO. At the 10th-percentile Gaussian discrepancy cutoff for source target-class recordings, full CFO achieved acceptance rates of 40.1\% for arousal and 58.4\% for valence. Our counterfactual explanations revealed that both models were sensitive to EEG window edges. Edits confined to the first and last 250 ms of each six-second window preserved all 202 tested prediction flips across valence and arousal.
\end{abstract}

\section{Introduction}
Electroencephalography (EEG) supports applications ranging from clinical and psychological assessment to brain-computer interfaces (BCIs) that infer cognitive, affective, or motor states. Neuroadaptive BCIs are applications that use estimated user states to modify an interface in real time, with potential benefits for adaptive games, education, and mixed-reality experiences \citep{ewing2016adaptive,kosmyna2019attentivu,bernal2022galea}. As wearable and consumer EEG systems become easier to integrate with everyday devices, EEG-based affect models may also support user experimentation informed by physiological measurements \citep{gasparfigueiredo2023eegux}. These applications motivate both accurate classification of affective states across participants and an understanding of how EEG features relate to model predictions. Counterfactual explanations (CFX), which identify the changes in features required to alter a given model's prediction, offer an approach to identify the relationship between EEG features and model predictions \citep{wachter2018counterfactual}. However, such changes must be physiologically plausible and compatible with the target-class distribution to support meaningful interpretation. This challenge is particularly relevant when models must generalize across subjects using noisy consumer-grade and sparse electrode setups \citep{katsigiannis2018dreamer}.

While a large body of work has applied deep learning (DL) architectures to publicly available EEG datasets for cognitive and affective state classification, comparatively less work has addressed the interpretability of these models. In this paper, we use leave-one-subject-out training and evaluation to study generalization to unseen participants \citep{khan2025reviewprocessfailures,delpup2025morebetter}. Efforts to improve predictive performance in these settings have produced increasingly complex architectures \citep{liu2021review,prabowo2023systematic}, making it difficult to determine which learned features and representations contribute to a model's decisions. Explainability methods can help characterize these predictive dependencies and inform how model outputs are used in downstream applications \citep{kindermans2019unreliability,ravindran2023empirical,pinto2023goal}. However, methods developed for other DL domains do not always transfer reliably to EEG because neural recordings are noisy, high-dimensional, and strongly influenced by preprocessing and feature construction \citep{ravindran2023empirical,khan2025reviewprocessfailures}. Gradient-based saliency methods are particularly susceptible to these issues because their attributions may reflect properties of the input representation or preprocessing pipeline rather than stable predictive dependencies learned by the model \citep{kindermans2019unreliability,haufe2014interpreting}. The reliability of backpropagation-based explanations also depends on how the explanation method propagates information through the network, further complicating their interpretation \citep{sixt2020explanations}.

These limitations motivate complementary explanation methods that characterize how changes in interpretable EEG features would alter a model's prediction. Counterfactual explanations provide a framework for examining such changes \citep{guidotti2024counterfactual} and aim to offer actionable insights \citep{mothilal2020dice, pinto2023goal}, with potential applications in neurofeedback and user experimentation. We investigate this approach through reconstructed latent representations learned for EEG valence and arousal prediction on DREAMER under leave-one-subject-out training and evaluation. Beyond prediction changes, we examine how counterfactual representations align with learned class distributions and explore subject-invariant typicality across the dataset.

The contributions of this study are listed as follows:
\begin{enumerate}
\item We jointly train a variational EEG classifier and waveform decoder
for valence and arousal. Subject-adversarial regularization discourages
the classification representation from encoding participant identity.

\item We propose a typicality-aware counterfactual optimizer for differentiable
encoder--decoder models trained with variational classification (VC)
\citep{dhuliawala2023variational}. The loss favors edits near the jointly learned target-class center, offering
an architecture-agnostic alternative to minimum-density constraints derived
from separately fitted Gaussian mixtures \citep{artelt2020convex}. It further
penalizes latent and waveform displacement and regularizes the amplitude,
band power, and inter-electrode connectivity of the decoded EEG.

\item On held-out DREAMER participants, we test whether target-class
predictions survive decoding and re-encoding. We use objective ablations,
a physiology-only repair control, and a comparison with adapted CoMTE to
assess trade-offs among decoded validity, target-class typicality,
proximity, and physiological plausibility.
\end{enumerate}

Typicality optimization aims to produce central counterfactual explanations for each class as represented by the model. We assess centrality by decoding the counterfactual and testing whether its Gaussian discrepancy is at or below the 10th percentile observed for real target-class recordings from the source participants while re-encoding the EEG to evaluate class-flip preservation.

To the best of our knowledge, prior work has not combined variational learning for EEG emotion prediction with typicality-aware counterfactual optimization (CFO) under subject-independent evaluation. We interpret the generated counterfactuals as model-level explanations and hypothesis-generating descriptions of electrode frequency and temporal patterns, rather than as direct evidence of causal changes in cortical generators. Establishing the latter would require additional neurophysiological and interventional validation.

Implementations of our DL architectures and counterfactual optimization method are released through the open-source \textit{EEGProc} library.

\section{Related Work}

\subsection{Deep Learning and EEG}


\paragraph{Autoencoding and Feature Fusion.}

Traditional EEG pipelines use handcrafted features such as band power, power spectral density, and entropy \citep{liu2021review}, whereas recent models learn spatio-temporal and spatio-spectral representations through autoencoding and feature fusion \citep{li2023stsnet,li2023mtlfusenet,hilali2025cmhfedan}. This motivates our jointly trained encoder--decoder architecture, which learns representations while preserving a reconstruction path for subsequent traversal.

\paragraph{Classification Targets.}\label{sec:vc}\label{sec:MLDG}

Meta-learning for Domain Generalization (MLDG) treats participants as separate domains and favors updates that generalize from meta-training to meta-test subjects \citep{li2018learning}. We use its first-order formulation without accessing data from the participant that is held out for evaluation. Then, VC combines discriminative prediction with class-conditional latent regularization, providing a probabilistic representation of class structure and uncertainty \citep{dhuliawala2023variational}. While \citet{dhuliawala2023variational} suggest using adversarial learning for discrimination, we model the class-conditional distributions of classification embeddings as diagonal Gaussians (see Appendix~\ref{app:adapted-vc}). Later, we decompose the classification target for our counterfactual target to distinguish target-class validity from distributional typicality (see Section~\ref{sec:distro_typ}). Additionally, focal loss may downweight easily classified observations and direct optimization toward difficult or ambiguous examples \citep{lin2017focal, li2023mtlfusenet}. Unlike \citet{li2023mtlfusenet}, we omit the class-weighting term $\alpha$ because our episode sampling supplies equal numbers of windows from each class.

\paragraph{Subject-Invariant Representation Learning.}\label{sec:sub_inv}\label{sec:sup}

We use \textbf{subject-invariance} to describe the aim of learning emotion-relevant representations with reduced participant-specific structure. Operationally, we apply subject-adversarial training to encourage reduced recoverability, but neither the training objective nor a low score from one subject head establishes complete independence from subject-specific structure. Subject-adversarial learning discourages representations from retaining participant-identifying information that may reflect individual spectral profiles, recording conditions, or artifacts rather than generalizable affective structure \citep{ozdenizci2020learning,delpup2025morebetter}. Our inverse subject adversarial head applies this principle to the window-level classification embedding through gradient reversal \citep{ganin2016domain}. We complement this objective with supervised contrastive learning, which draws same-class embeddings together while separating different classes \citep{khosla2020supervised}. We select positive pairs that share an emotion label but originate from different 
participants. Together with MLDG, these objectives organize the latent space around affective structure that transfers across subjects and is subsequently traversed by our counterfactual method.

\subsection{Explainability in Deep Learning}
DL models often achieve strong predictive performance at the cost of transparency, making it difficult to determine which aspects of the input drive a given prediction \citep{kindermans2019unreliability}. Explainability methods aim to address this problem by producing descriptions of model behavior. In practice, two major families are receiving attention in EEG predictive modeling: \textbf{saliency-based} methods, which assign importance scores to input features for a fixed prediction \citep{simonyan2013deep,sundararajan2017axiomatic}, and \textbf{counterfactual} methods, which search for a nearby alternative input that would change the model output \citep{wachter2018counterfactual,mothilal2020dice}. Saliency-based methods are useful for identifying which regions, channels, or features influence a prediction, while counterfactual methods are useful for characterizing how the input would need to change in order to obtain a desired prediction. This makes CFX particularly interesting for psychological diagnostics where it may show desired changes, motor imagery where it may map brain to motor divergences \citep{yin2024pioneering}, user experimentation where it may guide design decisions on iterations, and neurofeedback where it may find the desired EEG patterns for approximation.

\paragraph{Counterfactual Optimization.}
\citet{wachter2018counterfactual} formulate CFO as the search for the smallest change to an input $x$ that produces a desired output $y_{\mathrm{o}}$. While their proposed target balances validity and proximity \citep{guidotti2024counterfactual}, these metrics alone do not guarantee a useful counterfactual. Unconstrained optimization can produce off-manifold, unjustified, or unstable examples \citep{laugel2019dangers,jiang2024robust}. Practical formulations consequently augment \citet{wachter2018counterfactual}'s counterfactual target with regularizers for properties such as diversity, sparsity, plausibility, robustness, and temporal coherence \citep{mothilal2020dice,wang2024glacier,sevellec2025plausible,refoyo2026multispace,kostrzewa2026plausibility}. 

\citet{artelt2020convex} enforce plausibility through thresholded class-conditional density estimates, deriving convex quadratic constraints from Gaussian-mixture components. \citet{joshi2019revise} propose REVISE, which searches the latent space of a separately trained generative model and evaluates predictions on decoded candidates. Our framework builds on these established approaches in three respects. First, typicality uses the variational classifier's own learned class-conditional Gaussian, evaluated in the representation used for prediction with a threshold calibrated on each model's source participants. Second, the decoder is trained jointly with the subject-adversarial EEG encoder, so latent search and decoded explanations share a learned representation. Third, decoded full trials receive EEG-specific constraints on connectivity, amplitude, and band power. Decoded candidates are re-encoded and their window probabilities averaged to assess trial-level validity.

Lastly, \citet{lim2025ranking} define counterfactual typicality as the proportion of observations from the target class contained within a hyperball centered on the counterfactual, providing a proximity measure instead of distributional typicality.

\paragraph{Counterfactuals in EEG Modeling.}
CFX remains relatively uncommon in EEG, but the literature spans different tasks and representations. \citet{pinto2023goal} generated feature-level counterfactuals for seizure prediction by identifying minimal feature changes that reversed a prediction. \citet{yin2024pioneering} replaced time--frequency patches with features from an alternative motor-imagery class until the classifier decision changed. More recent work has used causal counterfactual analysis to distinguish causal from associative EEG patterns in autism evaluation using adversarial generative CFO \citep{wang2026counterfactual}. Another study modeled seizure recordings as spatio-temporal graphs to obtain plausible and temporally robust explanations \citep{zannotti2026cortex}. With respect to affective modeling, \citet{zhu2026causalcem} used counterfactual training combined with causal discovery to enhance multimodal emotion recognition. \citet{zhu2026causalcem} also provide a counterfactual physiological constraint that is methodologically different from the one we propose in Section~\ref{sec:CFO}.

\section{Methodology}
\subsection{Dataset Details and Data Preparation}
We evaluate the proposed framework on DREAMER, an affective-computing dataset comprising 128~Hz EEG recordings from 23 participants viewing 18 emotion-eliciting audiovisual stimuli with a 14-channel Emotiv EPOC headset \citep{katsigiannis2018dreamer}. After each stimulus, participants rated valence, arousal, and dominance on a five-point scale. We retain valence and arousal and define separate binary tasks, assigning ratings $\geq 3$ to the high class and ratings $<3$ to the low class. DREAMER supports leave-one-subject-out cross-validation (LOSOCV) and repeated CFO, while its extensive usage provides a prediction benchmark \citep{liu2021review, prabowo2023systematic, hilali2025cmhfedan, hua2026dualbranch, zhang2026rgbstyle}. It also utilizes a sparse, consumer-grade EEG headset that is typically more susceptible to noise, placement, and movements, making it ideal for uncertainty estimation. Finally, emotion recognition is a demanding test of typicality-aware explanation because affective states are abstract, and self-reported labels are subjective and highly variable across subjects \citep{ozdenizci2020learning, delpup2025morebetter}.

For each subject and trial, samples are chronologically ordered and checked for continuity, finite values, and label consistency. The 14 channels are detrended, notch-filtered at 50~Hz, and decomposed with fourth-order zero-phase Butterworth filters into $\theta$ ($4$--$8$~Hz), $\alpha$ ($8$--$13$~Hz), and $\beta$ ($13$--$30$~Hz) waveforms. Filtering is applied to the full contiguous stimulus before extracting the central 60~s to reduce boundary artifacts, yielding 7,680 samples per channel at 128~Hz. Treating each electrode--band pair as a separate input produces 42 channel-major features and an EEG tensor of size $23 \times 18 \times 42 \times 7{,}680$. Each recording supplies 10 non-overlapping six-second windows, each shaped $768\times42$ and labelled by the recording's rating. Subject and recording identities support sampling and evaluation but are not predictor inputs. Each window is divided by its global root-mean-square (RMS) amplitude across all 768 samples and 42 features, using a floor of $10^{-6}$. We retain this input-specific scale for physical-unit physiological measurements; generated outputs are not renormalized before reclassification.

\subsection{Cross-Validation Scheme}
We use 23-fold LOSOCV for valence and arousal. In each fold, we train a separate window model for each task from scratch on the other 22 participants and fit the counterfactual reference distributions using only their data \citep{khan2025reviewprocessfailures,delpup2025morebetter}. For emotion recognition, we freeze the epoch-50 classifier and average the ten window probabilities within each trial (Equation~\ref{eq:trial_probability_mean}). For each task, we select a participant-specific oracle threshold that maximizes balanced accuracy on that held-out participant's 18 labeled trials. Because class balance varies substantially across DREAMER participants (Appendix~\ref{app:trial_subject_results}) and participants may apply different criteria when rating their emotions, we use the participant-specific oracle results to help interpret the classifier and its counterfactuals on the observed trials. We also report performance at a fixed threshold. For counterfactual evaluation, both methods use the same frozen checkpoint for each fold. Eligibility uses the class with the highest mean probability across the ten windows. Both methods use a counterfactual confidence threshold of 0.65.

Training uses eight MLDG episodes per epoch. Each episode selects ten meta-training and five disjoint meta-test participants from the source pool. Each participant contributes one low-rated and one high-rated recording, including all 10 windows from each. This yields exactly class-balanced groups of 200 meta-training and 100 meta-test windows. Participants omitted from an episode remain eligible for later episodes. First-order MLDG takes a temporary inner u pdate on the meta-training classification, reconstruction, and subject-adversarial objective, evaluates the classification objective on meta-test participants, and applies one combined outer update. Held-out windows never enter these gradient updates. The counterfactual suite described in Appendix~\ref{app:cfo_window_suite} uses the frozen epoch-50 checkpoint.

\begin{figure}[t]
\centering
\includegraphics[width=0.9\columnwidth]{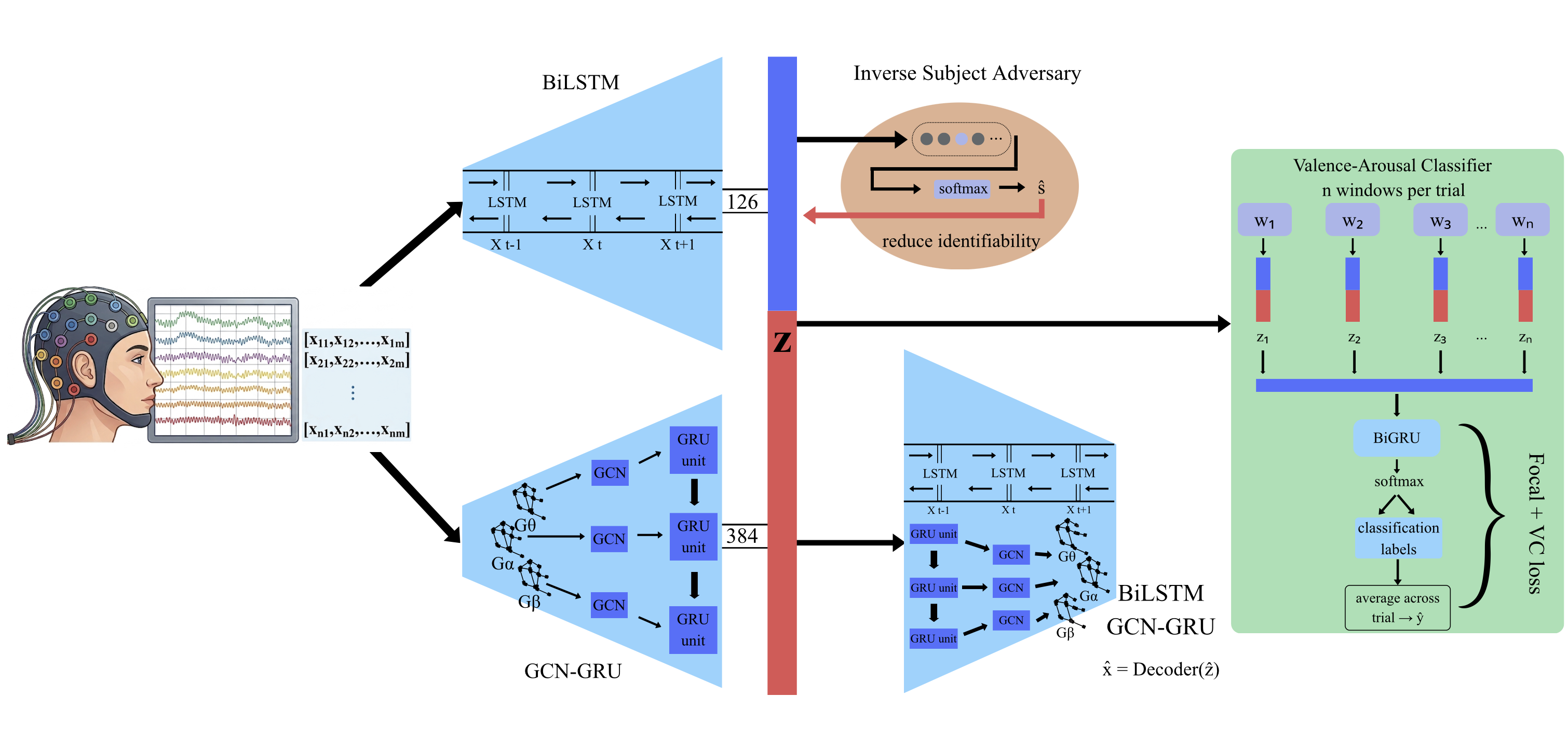}
\caption{Deep learning model workflow. Each six-second window is encoded independently. The BiGRU processes the 768 encoded timesteps within that window for affect prediction, while the decoder reconstructs channel--band waveforms and enables counterfactual traversal. Trial predictions average the probabilities of all ten windows. MLDG organizes source-participant training episodes.}
\label{fig:dl_workflow}
\end{figure}

\subsection{Spatio-Spectral-Temporal Encoding to Multi-Head Architecture}\label{sec:jmlo}

As illustrated in Fig.~\ref{fig:dl_workflow}, each input $\mathbf X_{s,t,w}\in\mathbb R^{768\times42}$ is one six-second window, indexed by participant $s$, recording $t$, and window $w$. A Bidirectional Long Short-Term Memory (BiLSTM) encoder learns temporal features, while a Graph Convolutional Network--Gated Recurrent Unit (GCN--GRU) encoder learns spatio-spectral features, inspired by \citet{li2023mtlfusenet}. Their fused output is a latent sequence $\mathbf Z_{s,t,w}\in\mathbb R^{768\times m}$, where $m$ is the fused feature dimension. A classifier with two bidirectional gated recurrent unit (BiGRU) layers, with 128 and 64 units respectively, processes the 768 latent timesteps within that window. Its final forward and backward states form the classification embedding supplied to the variational head, producing one high-valence or high-arousal probability per window. Windows are classified independently, and the ten window probabilities are averaged to obtain the trial prediction used for evaluation.

The decoder, classifier, and ISA are optimized jointly according to
\begin{equation}
\mathcal{L}_{\mathrm{joint}} =
\big(\lambda_{\mathrm{AE}}\mathcal{L}_{\mathrm{AE}} + \lambda_{\mathrm{sup}}\mathcal{L}_{\mathrm{sup}}\big)
+
\big(\lambda_{\mathrm{focal}}\mathcal{L}_{\mathrm{focal}}
+
\lambda_{\mathrm{VC}}\mathcal{L}_{\mathrm{VC}}\big)
+
\lambda_{\mathrm{ISA}}\mathcal{L}_{\mathrm{ISA}},
\end{equation}

where $\mathcal{L}_{\mathrm{AE}}$ measures autoencoder reconstruction error, $\mathcal{L}_{\mathrm{sup}}$ encourages cross-participant same-class alignment (Section~\ref{sec:sup}), $\mathcal{L}_{\mathrm{focal}}$ and $\mathcal{L}_{\mathrm{VC}}$ promote discriminative window-level predictions, $\mathcal{L}_{\mathrm{ISA}}$ discourages participant-identifying information, and $\lambda$ terms refer to non-negative weight hyperparameters. We use focal-loss exponent $\gamma=1.0$, reconstruction weight $\lambda_{\mathrm{AE}}=0.6$, and variational-classification weight $\lambda_{\mathrm{VC}}=1.0$. The reconstruction objective includes a seam penalty that matches the observed sample difference across chronologically adjacent windows and a derivative penalty that matches sample differences within each window; both temporal penalties have weight $1.0$. Joint optimization encourages embeddings that are sufficiently structured for reconstruction, predictive of affective state, and less informative about participant identity.

The VC regularizer may be decomposed as $\mathcal{L}_{\mathrm{VC}}
=
\beta\mathcal{L}_{\mathrm{posterior}}
+
\lambda\mathcal{L}_{\mathrm{prior}}$, where
\begin{equation}
\begin{aligned}
&\mathcal{L}_{\mathrm{posterior}} =
\frac{1}{d}\mathbb{E}_{p(y)}
\left[
D_{\mathrm{KL}}
\left(
q_{\phi}(z\mid y)
\,\middle\|\,
p_{\theta}(z\mid y)
\right)
\right]
=
\\[1mm]
&-\frac{1}{d}\mathbb{E}_{p(y)q_{\phi}(z\mid y)}
\left[
\log p_{\theta}(z\mid y)
\right]
-\frac{1}{d}\mathbb{E}_{p(y)}
\left[
\mathcal{H}\left[q_{\phi}(z\mid y)\right]
\right].
\end{aligned}
\label{eq:vc_reg}
\end{equation}
Here, $d$ is the classification-embedding dimensionality, and weights $\beta$ and $\lambda$ are non-negative. We set $\beta=1.0$ for valence and $\beta=0.5$ for arousal. The posterior term is averaged across embedding dimensions, matching the implementation and the closed-form objective in Appendix~\ref{app:adapted-vc}. This term aligns the empirical class-conditional latent
distribution with the distribution anticipated by the
classifier (see Appendix~\ref{app:adapted-vc}) and supplies a reference for subsequent typicality analysis. The expected negative log-density term favors embeddings supported by $p_\theta$, while the negative-entropy contribution discourages collapse of within-class variation. The remaining term, $\mathcal{L}_{\mathrm{prior}} = D_{\mathrm{KL}}(p(y)\|p_{\pi}(y))$, aligns the empirical label frequency with the learned class prior. The target classifier estimates
\begin{equation}
\hat p_{s,t,w}=p_\theta\!\left(y_{s,t,w}=1\mid\mathbf Z_{s,t,w}\right).
\end{equation}
Here, $y_{s,t,w}\in\{0,1\}$ denotes low or high valence or arousal, inherited from the recording's self-report. Recording labels supervise each window; no window-specific affect annotation is available. For each complete trial, prediction uses the arithmetic mean
\begin{equation}
\hat p_{s,t}=\frac{1}{W}\sum_{w=1}^{W}\hat p_{s,t,w},\qquad W=10.
\label{eq:trial_probability_mean}
\end{equation}
For emotion-recognition evaluation, we predict the high class when its trial-average probability meets the applicable fixed or participant-specific oracle threshold. A trial is eligible for counterfactual analysis if the class with the highest mean window probability matches the participant's recorded label. Exact ties are assigned to the low class. Thus, classification training uses windows, whereas the reported affective evaluation uses complete trials.

\subsection{Constrained Counterfactual Optimization}\label{sec:CFO}
In our architecture, optimization is performed in latent space, and a jointly trained decoder maps the optimized representations into reconstructed EEG features for interpretation. The joint architecture preserves a decoder for interpreting optimized latent sequences. Appendix~\ref{app:proof_of_sub_inv} states the limits of a subject-identifiability bound when physical-unit evaluation also uses input-specific RMS scaling. The CFO process described in Equation~\ref{eq:counterfactual-objective} moves a latent representation toward a target class probability, after which decoder traversal maps the optimized latent point into reconstructed EEG channel--band waveform space for interpretation. Therefore, reconstructed counterfactual feature representations must be measured both for domain-wise fidelity and for physiological realism, motivating the integration of validity, faithfulness, and plausibility constraints \citep{mothilal2020dice,wang2024glacier,sevellec2025plausible}.

Let $\mathbf Z=(\mathbf Z_1,\ldots,\mathbf Z_W)\in\mathbb R^{W\times T\times m}$ denote the complete trial's encoded windows, with $W=10$ and $T=768$. The frozen BiGRU summarizes each window as
\begin{equation}
C_\theta(\mathbf Z_w)
=\left[\overrightarrow{\mathbf h}_{w,T};\overleftarrow{\mathbf h}_{w,1}\right]
\in\mathbb R^d.
\end{equation}
The variational head maps each embedding to $f_c^{\mathrm{win}}(\mathbf Z_w)$. The full-trial classifier is $f_c(\mathbf Z)=W^{-1}\sum_{w=1}^{W}f_c^{\mathrm{win}}(\mathbf Z_w)$, matching Equation~\ref{eq:trial_probability_mean}. CFO jointly updates the complete collection of latent windows and targets this aggregated trial probability; it does not optimize each window as a separate counterfactual.

For binary valence or arousal classification, the counterfactual target is defined as the class opposite to the model's original prediction. Starting from the latent representation $\mathbf{Z}$, gradient-based optimization produces a counterfactual sequence $\mathbf{Z}^{\mathrm{cf}}$ that increases the probability of the target class while limiting its deviation from the original representation:
\begin{equation}
\begin{aligned}
\mathcal{L}_{\mathrm{cfo}}
=
&\lambda_o\mathcal{L}_{\mathrm{o}}\left(f(\mathbf{Z'}),y_{\mathrm{o}}\right)
+\lambda_zd_z(\mathbf{Z},\mathbf{Z'})
+\lambda_Rd_R\left(R(\mathbf{Z}),R(\mathbf{Z'})\right)
\\
&+\lambda_{\mathrm{ph}}\left[\mathcal P_{\mathrm{ph}}\!\left(R(\mathbf Z')\right)
+\mathcal P_{\mathrm{AP}}\!\left(\mathbf r\odot R(\mathbf Z')\right)\right],
&\mathbf{Z}^{\mathrm{cf}}
=
\underset{\mathbf{Z'}}{\arg\min}
~\mathcal{L}_{\mathrm{cfo}}.
\label{eq:counterfactual-objective}
\end{aligned}
\end{equation}
In Equation~\ref{eq:counterfactual-objective}, $\mathbf Z$ denotes the original full-trial latent representation, $\mathbf Z'$ the jointly optimized candidate, and $\mathbf Z^{\mathrm{cf}}$ the selected counterfactual. The fixed classifier $f$ averages the window probabilities, while $R$ reconstructs all windows. The fixed vector $\mathbf r=(r_1,\ldots,r_W)$ contains the original windows' global RMS amplitudes; $\mathbf r\odot R(\mathbf Z')$ broadcasts each scale over its corresponding window to restore physical units. The target loss promotes $y_{\mathrm o}$, the two distances penalize latent and reconstructed displacement, $\mathcal P_{\mathrm{ph}}$ denotes VCSC, and $\mathcal P_{\mathrm{AP}}$ penalizes amplitude/band-power deviations. The two physiological penalties have distinct definitions and share the enabled physiological weight in this suite. Only $\mathbf Z'$ is updated; model weights, calibration references, and RMS scales stay fixed.

\paragraph{Validity and Decoder Faithfulness.}
For each correctly classified trial, we assess target assignment and confidence
using the mean class probabilities across its ten windows. In latent space,
target assignment requires $\arg\max_c f_c(\mathbf Z^{\mathrm{cf}})=y_{\mathrm o}$,
and confident assignment additionally requires
$f_{y_{\mathrm o}}(\mathbf Z^{\mathrm{cf}})\geq 0.65$ for both outcomes and
directions. We apply the same tests after re-encoding the normalized
counterfactual $\widetilde{\mathbf X}^{\mathrm{cf}}=R(\mathbf Z^{\mathrm{cf}})$
with the frozen encoder $g_\phi$. We also reclassify the unedited reconstruction
$R(\mathbf Z)$. A newly induced decoded flip occurs when the reconstruction
retains the original class and the decoded counterfactual receives the target
class.

Before counterfactual intervention, we report the joint decoder's reconstruction $R^2$ on
RMS-normalized EEG, pooling all input values within each held-out participant
and averaging the participant-level scores. For each counterfactual, we measure
waveform displacement by RMSE across all ten normalized windows.
$E_{\mathrm{cf}}=\operatorname{RMSE}(\widetilde{\mathbf X}^{\mathrm{cf}},\mathbf X)$
measures total displacement from the observed input, and
$\Delta_{\mathrm{dec}}=\operatorname{RMSE}(\widetilde{\mathbf X}^{\mathrm{cf}},R(\mathbf Z))$
measures the edit relative to the unedited reconstruction. For the matched
repair-only control, we also report RMSE between the reconstruction and repair
and between the repair and counterfactual endpoints. Physical-unit descriptors
use $\mathbf X^{\mathrm{cf}}=\mathbf r\odot\widetilde{\mathbf X}^{\mathrm{cf}}$
with the original per-window RMS scales.

\paragraph{Physiological Plausibility.}
We calibrate both physiological penalties on real trials from the training
participants in each fold. The volume-conduction spatial coherence (VCSC)
penalty $\mathcal P_{\mathrm{ph}}$ measures deviations in magnitude-squared
coherence and debiased squared weighted phase-lag index (wPLI) across
electrode pairs. The amplitude and band-power penalty $\mathcal P_{\mathrm{AP}}$
penalizes physical-unit measurements outside their central 95\% reference
intervals. To evaluate each decoded trial, we report a physiology grade
$G=1/(1+\bar e_{\max})$, where $\bar e_{\max}$ is the largest mean
exceedance across the scorable families. Higher grades indicate closer
agreement with the reference ranges. Appendices~\ref{app:vcsc}
and~\ref{app:graded_physiology} give the estimators and scoring rules.

\subsubsection{Distributional Typicality}
\label{sec:distro_typ}

For either target class $c\in\{0,1\}$, the frozen variational head defines a diagonal Gaussian in the $d$-dimensional window-classification embedding space:
\begin{equation}
p_{C_c}(u)=p_\theta(u\mid y=c)
=\mathcal N\!\left(u\mid\mu_{C_c},
\operatorname{diag}(\sigma_{C_c}^{2})\right),
\qquad u\in\mathbb R^d.
\end{equation}
The Gaussian parameters are copied from the trained checkpoint. With variance floor $\varepsilon>0$, the complete trial's discrepancy averages the window scores:
\begin{equation}
D_{C_c}(\mathbf Z)
=\frac{1}{Wd}\sum_{w=1}^{W}\sum_{j=1}^{d}
\frac{\left(C_{\theta,j}(\mathbf Z_w)-\mu_{C_c,j}\right)^2}
{\max(\sigma_{C_c,j}^{2},\varepsilon)}.
\label{eq:class1_latent_discrepancy}
\end{equation}
For each model and target class, $\tau_c$ is the empirical 10th percentile of complete source-trial discrepancies with true class $c$. Gaussian acceptance is $\mathcal T_{C_c}=\{\mathbf Z:D_{C_c}(\mathbf Z)\leq\tau_c\}$. The cutoff applies to the mean trial score and does not require each constituent window to pass an individual cutoff. This strict criterion targets a narrow central region under the shared class score. Acceptance measures how often held-out counterfactuals reach this low-discrepancy region. Section~\ref{sec:results} reports source-training accuracy for the trained subject-identification head. The evaluated participant contributes neither to Gaussian fitting nor to cutoff calibration. Table~\ref{tab:gaussian_acceptance} reports Gaussian acceptance for the generated counterfactuals and the repair control.

Joint latent confidence and Gaussian acceptance require
\begin{equation}
f_c(\mathbf Z^{\mathrm{cf}})\geq p_{\mathrm o}
\quad\wedge\quad
D_{C_c}(\mathbf Z^{\mathrm{cf}})\leq\tau_c.
\label{eq:cfo_condition}
\end{equation}
In the evaluated epoch-50 suite, the optimizer decodes the candidate latent sequence, re-encodes the resulting EEG with the frozen encoder $g_\phi$, and scores the resulting representation. When the typicality term is active, its objective is
\begin{equation}
\mathcal L_{\mathrm{cfo}}^{\mathcal T_{C_c}}
=\mathcal L_{\mathrm{cfo}}
+\lambda_{\mathcal T}\,
\frac{D_{C_c}\!\left(g_\phi(R(\mathbf Z'))\right)}{\max(\tau_c,\varepsilon)}.
\label{eq:typicality_cfo_objective}
\end{equation}
The typicality stage activates after the trial target assignment and confidence are reached and, when enabled, VCSC is feasible. Once activated, it remains active within the available budget. Amplitude/band-power penalties remain in the objective and in final feasibility, but do not delay typicality activation. Normalization by $\tau_c$ scales the penalty without changing the acceptance threshold. The penalty remains active inside the accepted region.

At the selected endpoint, Gaussian acceptance applies the same discrepancy $D_{C_c}(g_\phi(R(\mathbf Z^{\mathrm{cf}})))$ to the delivered waveform. For the epoch-50 suite, joint decoded success requires decoded target confidence, decoded Gaussian acceptance, and passage of both the amplitude and band-power checks. The broader physiological assessment also evaluates connectivity and VCSC feasibility. All success rates retain every eligible attempt, including optimization failures.

\subsection{Counterfactual Experiments and Ablations}
\label{sec:cfo_experiments}

We evaluate the window classifier--decoder across all 23 LOSOCV folds for valence and arousal. Detailed protocols appear in Appendix~\ref{app:cfo_experimental_details}.

\paragraph{Full-Trial Counterfactual Suite.}
\label{sec:cfo_window_suite}

Within each evaluated LOSOCV fold, we target the opposite class on correctly classified complete trials by jointly optimizing their ten six-second windows. We compare six conditions: (i) target loss with latent proximity, $\mathcal L_{\mathrm{base}}$; (ii) CFO adding decoded proximity and physiology, $\mathcal L_{\mathrm{cfo}}$; (iii) full CFO adding typicality, $\mathcal L_{\mathrm{cfo}}^{\mathcal T_{C_c}}$; (iv) full CFO without physiology, $\mathcal L_{\mathrm{cfo}}^{\mathcal T_{C_c}-\lambda_{\mathrm{ph}}}$; (v) full CFO with decoded proximity as its only proximity constraint, $\mathcal L_{\mathrm{cfo}}^{\mathcal T_{C_c}-\lambda_z}$, setting latent proximity to zero while retaining the other full-objective terms; and (vi) physiology-only repair, $\mathcal L_{\mathrm{repair}}$, retaining proximity but removing target and typicality terms. These symbols follow the definitions in Appendix~\ref{app:cfo_window_suite}. All conditions share frozen models, initialization, seeds, and the maximum optimization budget. Base CFO stops at its first feasible candidate. Full CFO can continue refining and selects the feasible candidate with the lowest Gaussian discrepancy. The comparison therefore measures the combined effect of the penalty and these search rules. Decoded signals are re-encoded to assess target validity and Gaussian acceptance. Full-trial physiological checks cover amplitude, band power, coherence, debiased squared wPLI, and VCSC feasibility. The repair control separates class-targeted changes from physiological correction.

\paragraph{Comparison with CoMTE.}
\label{sec:cfo_comte} We compare adapted CoMTE \citep{ates2021comte} with CFO using the same trial-averaged classifier, eligible complete trials, targets, and endpoint metrics. CoMTE substitutes electrode groups from correctly classified source target-class donors. Its direct use of real donor waveforms avoids decoder reconstruction error and may favor amplitude/band-power scores, while its restricted donor substitutions limit the changes it can express. Failed attempts remain in both methods' denominators; Appendix~\ref{app:cfo_comte} details these tradeoffs and the method-specific budgets.

\section{Results}
\label{sec:results}

\subsection{Emotion Recognition on DREAMER}
\label{sec:emotion_results}

At participant-specific oracle thresholds, mean trial-level balanced accuracy (BA) was 65.28\% for valence and 64.83\% for arousal. At the fixed threshold, mean BA was 55.32\% for valence and 52.16\% for arousal. At the oracle thresholds, both models had higher recall for low-rated trials, despite high ratings being more common. Both decoders explained approximately 47\% of the variance in RMS-normalized EEG features (Table~\ref{tab:emotion_recognition}).

\begin{table}[H]
\centering
\caption{Trial-level balanced accuracy (BA) and class recall at epoch 50, averaged across the 23 held-out DREAMER participants. Decoder $R^2$ measures EEG reconstruction.}
\label{tab:emotion_recognition}
\begin{tabular}{lccccc}
\hline
& \textbf{Fixed Threshold} & \multicolumn{3}{c}{\textbf{Oracle Threshold}} & \textbf{Decoder} \\
\textbf{Task} & \textbf{BA (\%)} & \textbf{BA (\%)} & \textbf{Recall 0 (\%)} & \textbf{Recall 1 (\%)} & $\boldsymbol{R^2}$ \\
\hline
Valence & 55.32 & 65.28 & 71.27 & 59.29 & 0.4747 \\
Arousal & 52.16 & 64.83 & 75.82 & 53.84 & 0.4698 \\
\hline
\end{tabular}
\end{table}

Across the 414 DREAMER trials, high ratings were more common for arousal (72.46\%) than for valence (61.11\%). At participant-specific oracle thresholds, median trial-level BA was 66.25\% for valence and 64.29\% for arousal. Across participants, BA ranged from 50.00\% to 76.92\% for valence and from 50.00\% to 82.35\% for arousal. Three participants had only one low-arousal trial each. Whether that trial was classified correctly determined their entire low-arousal recall, which accounts for half of BA (Appendix~\ref{app:model_report}).

\paragraph{Subject Identification With and Without Gradient Reversal.}
The trained subject-identification head identified participants with over 20\% accuracy when the gradient reversal layer (GRL) was disabled. With GRL enabled, the head's final source-training accuracy averaged 4.65\% across 23 LOSO folds, close to the 4.55\% expected from uniform guessing among the 22 source participants in each fold. The lower accuracy supports the subject-invariance objective because the trained head was less able to identify participants from the learned embeddings. The optimal identification accuracy across all possible predictors remains unknown (Appendix~\ref{app:proof_of_sub_inv}).

\subsection{Evaluation of Counterfactual Optimization}
\label{sec:cfo_results_draft}

We evaluated counterfactual optimization (CFO) on all 23 held-out participants for each task. All five class-targeted CFO objectives, the untargeted repair control, and CoMTE completed on 214 correctly classified valence recordings and 217 correctly classified arousal recordings. Each targeted run requested the opposite of the original prediction. Low-to-high targets accounted for 77/214 valence recordings (36.0\%) and 34/217 arousal recordings (15.7\%). We pool recordings across directions within each task unless otherwise specified. Appendix~\ref{app:cfo_experimental_details} describes the protocol, and Appendix~\ref{app:cfo_results} reports the results by condition.

\noindent\textbf{Prediction preservation before intervention.}
Decoding and re-encoding the unedited representation changed the original prediction on 81/214 valence recordings (37.9\%) and 53/217 arousal recordings (24.4\%). After physiological repair, which had no target-class objective, 92/214 valence predictions (43.0\%) and 47/217 arousal predictions (21.7\%) differed from their original class (Table~\ref{tab:prediction_preservation}). However, confident flips after physiological repair occurred for only 15/214 valence predictions (7.0\%) and 7/217 arousal predictions (3.2\%) (Table~\ref{tab:prediction_preservation}). We therefore also measured target hits on the 133 valence and 164 arousal recordings whose original prediction survived unedited reconstruction. On these recordings, reaching the target required an additional change beyond reconstruction alone.

\noindent\textbf{Latent and decoded target confidence.}
Every targeted objective reached a latent target probability of at least 0.65 on every eligible recording. After decoding and re-encoding, full CFO assigned the target class to 190/214 valence recordings (88.8\%) and 146/217 arousal recordings (67.3\%). Among recordings whose original prediction survived reconstruction, full CFO produced the requested flip on 109/133 (82.0\%) and 93/164 (56.7\%), respectively. The base objective without Gaussian typicality produced flips on 35/133 (26.3\%) and 15/164 (9.1\%). Across all eligible recordings, decoded confidence passage rose from 6.5\% with base CFO to 47.2\% with full CFO for valence, and from 3.7\% to 29.0\% for arousal (Table~\ref{tab:counterfactual results}).

\noindent\textbf{Target-class Gaussian acceptance.}
We assessed how centrally the model represented each re-encoded waveform by comparing its Gaussian discrepancy with the distribution for real target-class recordings from source participants. At the 10th-percentile cutoff (P10), acceptance rose from 11/214 (5.1\%) with base CFO to 125/214 (58.4\%) with full CFO for valence, and from 9/217 (4.1\%) to 87/217 (40.1\%) for arousal. Full CFO met both P10 and the 0.65 decoded confidence threshold on 72/214 valence recordings (33.6\%) and 42/217 arousal recordings (19.4\%). At the broader P25 cutoff, 92/214 (43.0\%) and 56/217 (25.8\%) met both criteria (Table~\ref{tab:gaussian_acceptance}). Joint decoded success as defined in Section~\ref{sec:distro_typ}, which also requires the amplitude and band-power checks, was reached by 17/214 valence and 15/217 arousal recordings under full CFO. These checks are calibrated on recorded EEG from the other participants (Appendix~\ref{app:graded_physiology}), and 199/214 and 203/217 of the recorded trials pass both. Under a flat rule requiring 95\% of components inside their intervals, only 148/214 and 138/217 recorded trials would pass, and full CFO would reach joint decoded success on 5/214 and 6/217 recordings.

\noindent\textbf{Physiological and proximity penalties.}
Removing the physiological penalties raised the fraction meeting both P10 and decoded confidence to 213/214 (99.5\%) for valence and 215/217 (99.1\%) for arousal. Mean physiology grades fell from 0.964 to 0.847 and from 0.957 to 0.835, respectively, where higher grades indicate closer agreement with physiological reference ranges. These edits had smaller mean waveform RMSE relative to the unedited reconstruction (0.098 versus 0.283 for valence and 0.101 versus 0.287 for arousal), and their grades equal those of the unedited reconstructions (0.847 and 0.835; per-trial mean difference below 0.001). The physiological deviation is therefore set by the decoder output rather than by the edit, and the physiological penalties raise the grade mainly by correcting the reconstruction: the untargeted repair control alone reaches 0.951 and 0.947. Removing latent proximity while retaining decoded proximity left target assignment and confidence counts unchanged for both tasks. One additional valence recording met both P10 and confidence, while the arousal count stayed the same (Tables~\ref{tab:counterfactual results} and~\ref{tab:gaussian_acceptance}).

\noindent\textbf{Comparison with CoMTE.}
Across all 23 held-out participants, full CFO and CoMTE reached the target class on 190/214 (88.8\%) and 188/214 (87.9\%) valence recordings, respectively, and on 146/217 (67.3\%) and 145/217 (66.8\%) arousal recordings. Among recordings whose original prediction survived unedited reconstruction, CoMTE produced more target hits on both tasks (114 versus 109 of 133 for valence and 98 versus 93 of 164 for arousal). Full CFO reached a target probability of at least 0.65 more often than CoMTE (47.2\% versus 31.3\% for valence and 29.0\% versus 15.2\% for arousal). Full CFO also met both confidence and P10 more often (72/214 versus 11/214 for valence and 42/217 versus 4/217 for arousal). Of the 414 original recordings, 120 were eligible for both tasks, yielding 431 matched task--trial pairs from 311 distinct recordings. Across these pairs, recorded generation time averaged 107.7~s per trial for full CFO and 160.3~s for CoMTE under the respective run conditions ($1.49\times$ as long for CoMTE). Full CFO stayed closer to the unedited reconstruction on average (Table~\ref{tab:counterfactual results}). Relative to the recorded EEG, CoMTE had lower mean waveform RMSE for both valence (0.650 versus 0.783 for full CFO) and arousal (0.654 versus 0.795).

\subsection{Single Subject-Trial Case Analysis}\label{sec:case_study}

We illustrate full CFO on a correctly classified high-arousal recording from held-out participant~19, trial~2, with low arousal as the target ($c=0$). Among the 24 of 183 high-to-low arousal recordings whose original prediction survived unedited reconstruction and whose decoded counterfactual met both target confidence and P10 acceptance, it is the second closest to the population medians on nine quantities, including $d_z$, $\Delta_{\mathrm{dec}}$, $E_{\mathrm{rec}}$, the decoded probability gain and the physiology grade (Appendix~\ref{app:case_subjects}). The latent counterfactual raised the low-arousal probability from 0.416 to 0.936 at $d_z=0.262$ (median 0.261 over the 183 recordings). After decoding and re-encoding, the unedited reconstruction $R(\mathbf Z)$ received 0.419 and the counterfactual 0.704, so the edit induced a confident decoded flip. Its re-encoded discrepancy fell from $D_{C_0}=1.070$ for $R(\mathbf Z)$ to $1.058$, below $\tau_0=1.066$. The counterfactual moved $\Delta_{\mathrm{dec}}=0.325$ from the reconstruction, against $E_{\mathrm{rec}}=0.764$ and $E_{\mathrm{cf}}=0.832$ relative to the recorded EEG.

\begin{figure}[h]
    \centering
    \includegraphics[width=0.9\linewidth]{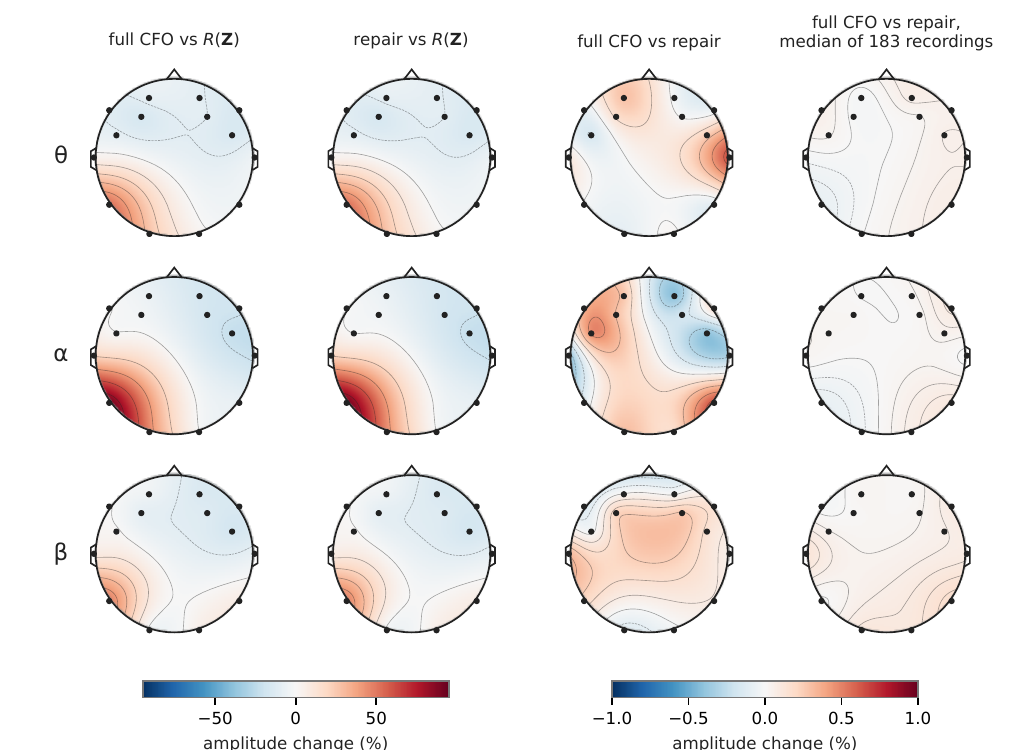}
    \caption{Band-amplitude change on the scalp for the arousal case recording (participant 19, trial 2). Amplitude is the 99th percentile of $|x|$ over the full trial in physical units, and maps interpolate the 14 electrodes with a thin-plate spline. First two columns: full-CFO counterfactual and repair endpoint relative to $R(\mathbf Z)$ (left colour scale). Last two columns: full CFO relative to repair for this recording, and the median of that difference over all 183 high-to-low arousal recordings (right colour scale, $\pm1\%$).}
    \label{fig:topomap_arousal}
\end{figure}

The physiology grade was 0.996 for the recorded trial, 0.793 for $R(\mathbf Z)$ and 0.990 for the counterfactual. The counterfactual was VCSC-feasible and passed the coherence (97.8\% of components within the reference interval), debiased squared wPLI (99.6\%) and amplitude (83.3\%) checks, but missed band power (73.8\% against a required 74.2\%) by one of 42 components, so it does not meet joint decoded success; across the suite, 15 of 217 arousal counterfactuals do. Its largest family mean exceedance, 0.010 half-interval widths, shows that the failing components lie just outside their intervals.

The other conditions run on the same recording separate the contributions of the terms. The repair control, without target or typicality terms, moved the decoded signal by the same amount ($\Delta_{\mathrm{dec}}=0.326$), reached the same grade (0.990) and left the re-encoded low-arousal probability at 0.491. $\mathcal L_{\mathrm{base}}$ flipped the latent prediction (0.717) but barely changed the decoded signal ($\Delta_{\mathrm{dec}}=0.002$), which re-encoded at 0.443. Without the physiological penalties, the counterfactual re-encoded at 0.9997 with $\Delta_{\mathrm{dec}}=0.111$, and its grade stayed at the reconstruction's 0.793. Band amplitudes do not separate the full counterfactual from the repair endpoint. We took the 99th percentile of $|x|$ over the trial for each electrode and band: the full counterfactual and the repair endpoint change these 42 values from $R(\mathbf Z)$ almost identically (correlation 0.9999; root-mean-square difference 0.25 percentage points against a root-mean-square change of 22.1), and their difference stays below 0.7\% at every electrode and band (Figure~\ref{fig:topomap_arousal}). The visible band-level change, dominated by an increase at P7 (91\% in alpha, 51\% in theta and 47\% in beta) and O1, therefore comes from the physiological penalties. Across the 183 high-to-low arousal recordings, the median difference between full CFO and repair is at most 0.17\%, and its sign varies between recordings (positive in 41--73\% of recordings at each electrode and band), so the population summary shows no consistent scalp pattern either.

\begin{figure}[h]
    \centering
    \includegraphics[width=0.49\linewidth]{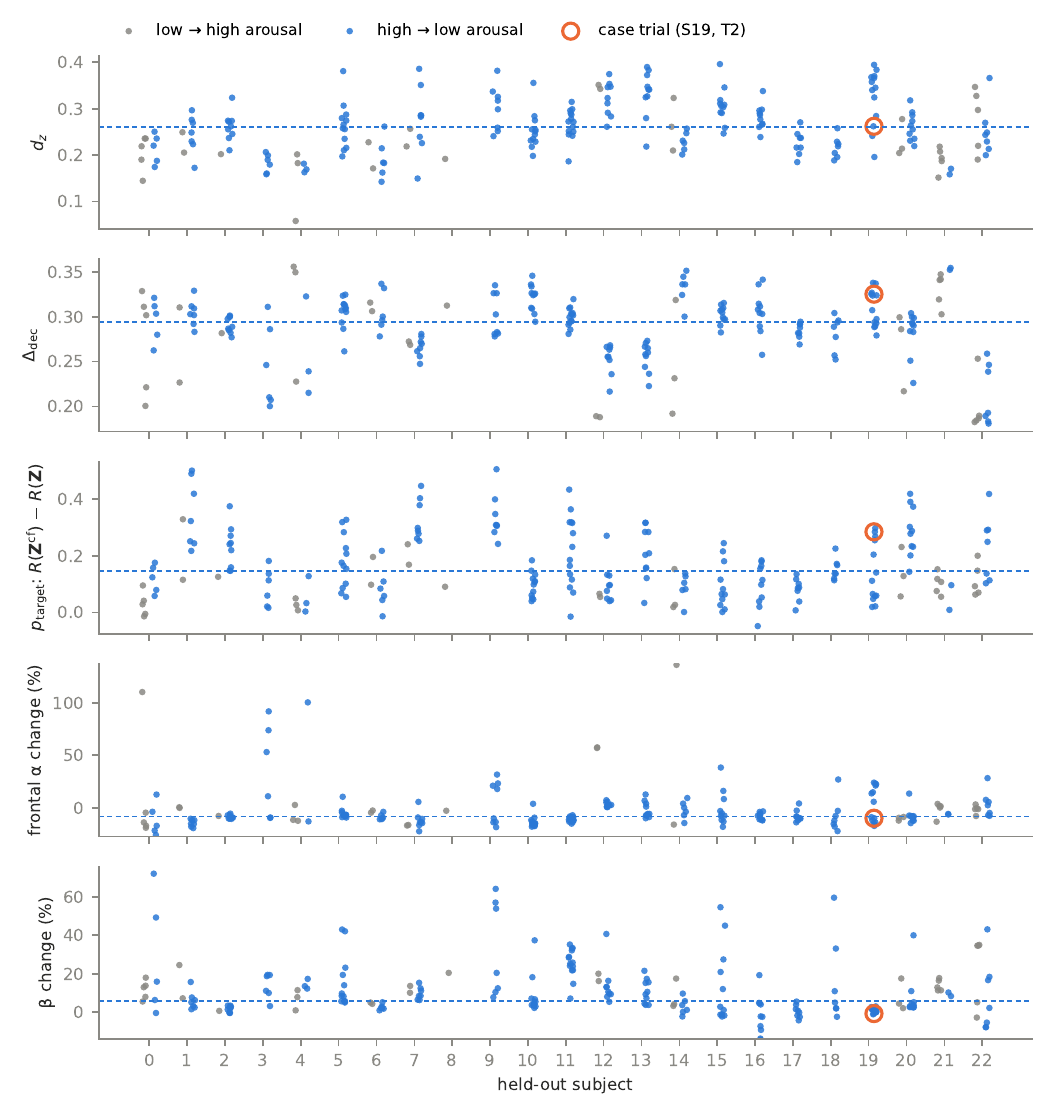}
    \hfill
    \includegraphics[width=0.49\linewidth]{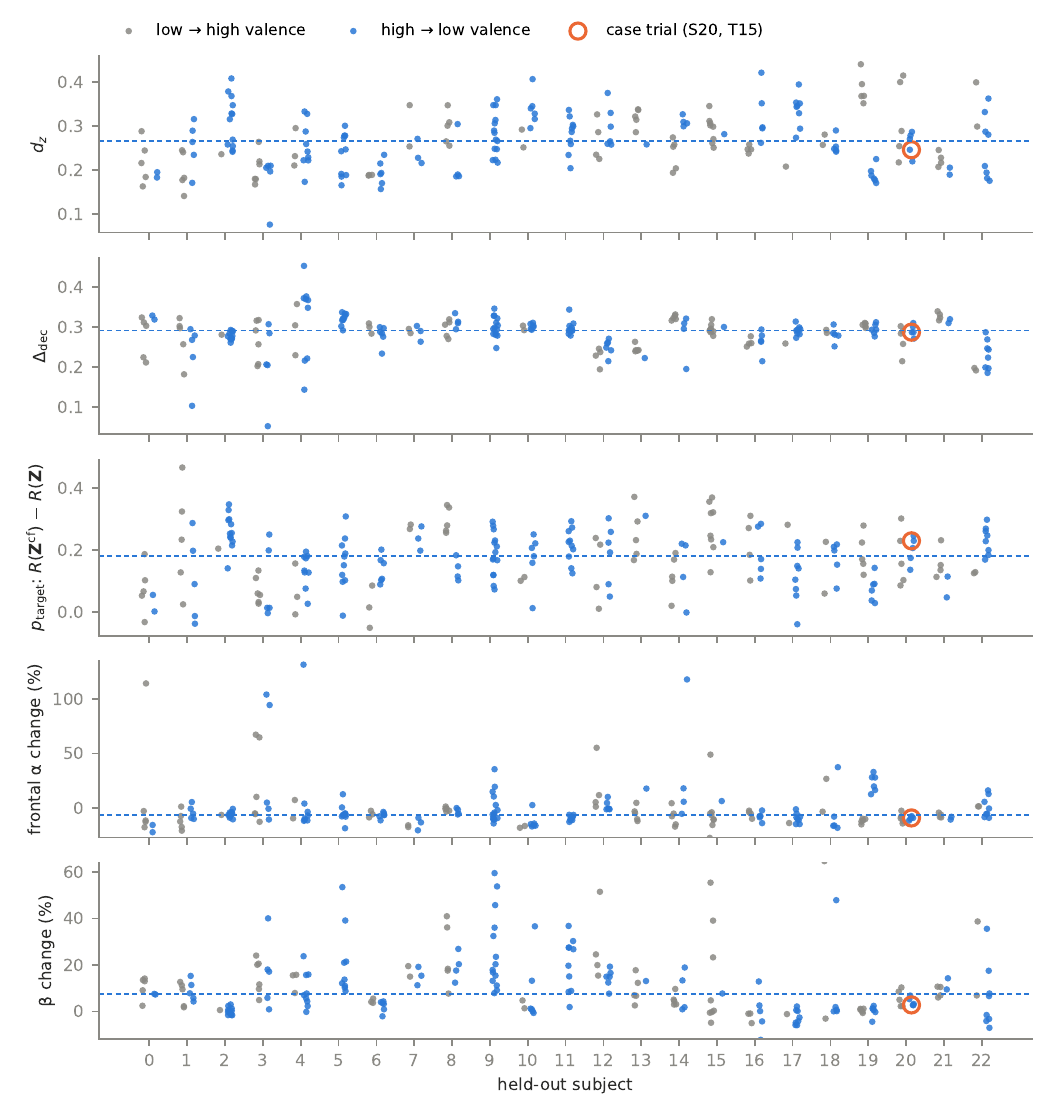}

        \caption{Case-study quantities for every eligible recording under full CFO, by held-out participant, for arousal (left) and valence (right). Blue: high-to-low targets; grey: low-to-high. The ringed point is the case recording; dashed lines mark high-to-low medians.}
    \label{fig:subject_comparison}
\end{figure}

To test which parts of each window drive the prediction, we subtract the repair endpoint from the full counterfactual and mask the resulting edit. In the median recording, 60\% (arousal) and 57\% (valence) of the edit energy lies in the first and last 0.25~s (32 samples) of each six-second window, which together make up 8\% of the samples. Without the physiological penalties, 85\% and 89\% of the edit energy lies in these edges. With each participant's frozen model, re-encoding the unmasked edits reproduced the stored decoded probabilities to within $3\times10^{-4}$. Retaining only the edges of the edit preserved the target-class predictions of full CFO on all 93 arousal and 109 valence recordings in the masking analysis. The counts reaching a target probability of 0.65 and passing P10 also stayed the same, and the case recording kept its target probability of 0.704. Keeping only the middle 92\% of each window yielded target-class predictions on 10 arousal and 29 valence recordings. None of those arousal recordings and one valence recording reached a target probability of 0.65. The case recording returned to the repair value of 0.491 (Table~\ref{tab:edge_mask}). Beta accounted for a median of 61\% of the edit energy for arousal and 64\% for valence.

The use of the BiGRU's terminal states offers a possible explanation for this boundary sensitivity. The window embedding uses the final forward and backward states. In a recorded-window probe on four participants per task, replacing the middle five seconds of each window with data from another recording of the same participant while retaining the first and last 0.5~s changed the window probability by a median of $3\times10^{-4}$ (arousal) and $1\times10^{-4}$ (valence), compared with 0.15 and 0.18 when the whole window was replaced (Appendix~\ref{app:case_subjects}). Whole-trial band summaries can obscure edits concentrated at the window boundaries.

Across participants, the case recording is typical in proximity and decoded change but belongs to a minority (Figure~\ref{fig:subject_comparison}). Of the 183 high-to-low arousal recordings, 35 met both decoded confidence and P10 acceptance, 24 of them with a flip induced by the edit, from 9 of 22 participants. Differences between participants account for 54\%, 66\% and 54\% of the variance in $d_z$, $\Delta_{\mathrm{dec}}$ and the decoded probability gain. For valence, 24 of 137 high-to-low recordings met both criteria with an induced flip, from 10 of 23 participants (Appendix~\ref{app:case_subjects}). We also looked for band or spatial patterns characteristic of the low classes and found none in the decoded signal. Without the physiological penalties, the median decoded changes toward the low and high classes show similar patterns (correlation 0.96 for arousal and 0.93 for valence, mainly an increase in posterior and temporal beta amplitude in both directions), and a leave-one-participant-out test cannot separate the two directions from amplitude, coherence or repair-adjusted amplitude changes (AUROC 0.32--0.50). We therefore do not interpret band-level changes as markers of the target class.

\subsection{Bandwave Continuity and Connectivity of Decoded Signals}\label{sec:continuity}

\begin{figure}[t]
    \centering
    \includegraphics[width=0.9\linewidth]{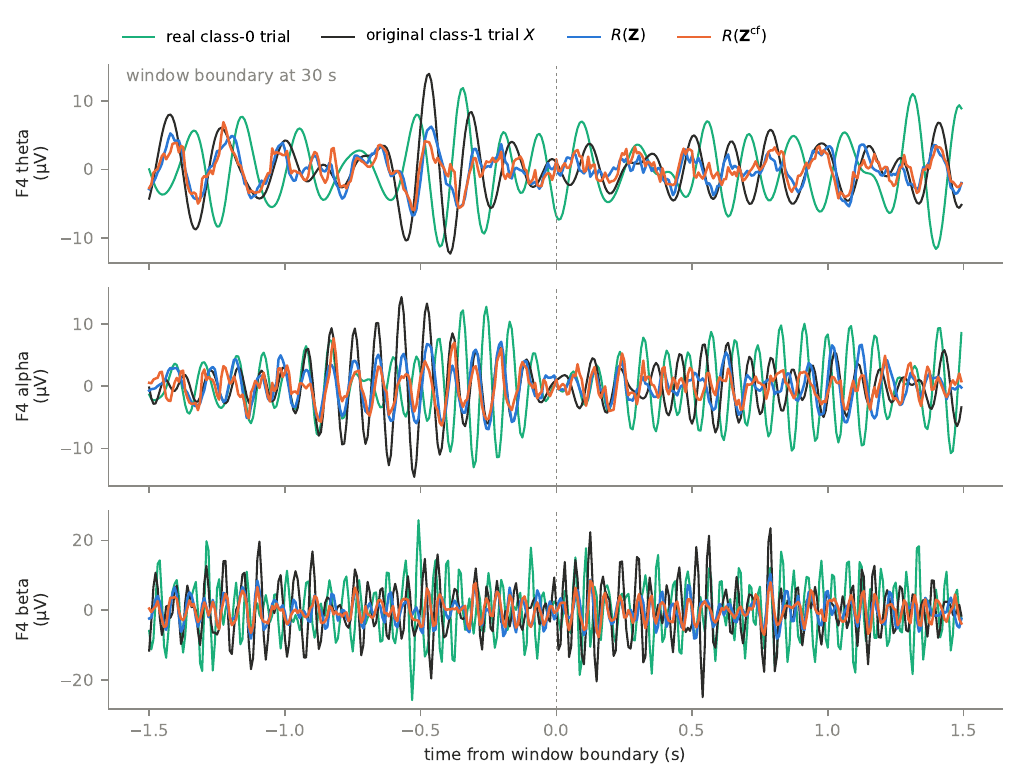}
    \caption{Electrode F4 around the boundary between windows 5 and 6 of the arousal case recording (participant 19, trial 2): the recorded high-arousal trial, its unedited reconstruction $R(\mathbf Z)$, the full-CFO counterfactual $R(\mathbf Z^{\mathrm{cf}})$, and a recorded low-arousal trial of the same participant (trial 1). Median boundary jumps for this recording: recorded 1.12, $R(\mathbf Z)$ 1.68, $R(\mathbf Z^{\mathrm{cf}})$ 1.97.}
    \label{fig:bandwave_waveform}
\end{figure}

The decoder reconstructs each six-second window separately, so a decoded recording can be discontinuous at the nine boundaries where consecutive windows meet. We measure the amplitude jump at each boundary, the sample-to-sample step across it relative to the steps around it, which is about 1 for continuous EEG (definitions and results by condition in Appendix~\ref{app:bandwave}). The recorded EEG gives a median jump of 1.04 in both tasks. The unedited reconstruction already breaks continuity (1.74 for valence and 1.80 for arousal), mostly in theta and alpha. Full CFO raises the jump to 2.02 and 2.07, and in 81\% (valence) and 82\% (arousal) of recordings the counterfactual has larger boundary jumps than its own reconstruction. The increase comes from the physiological penalties: the target and latent-proximity condition leaves the boundaries unchanged, and the repair control raises them at least as much as full CFO (2.09 and 2.13). The class-targeted edit adds to this at the same place: it concentrates at the window edges (Section~\ref{sec:case_study}), and without the physiological penalties it alone raises the jump from 1.74 to 1.89 (valence) and from 1.80 to 2.07 (arousal). VCSC and the amplitude/band-power penalty are computed over the whole recording and do not penalize discontinuities at window boundaries. Decoded counterfactuals are therefore continuous within windows but not across them, and we evaluate them per recording rather than as continuous signals. Figure~\ref{fig:bandwave_waveform} shows one boundary of the arousal case recording.

\begin{figure}[H]
    \centering
    \includegraphics[width=0.9\linewidth]{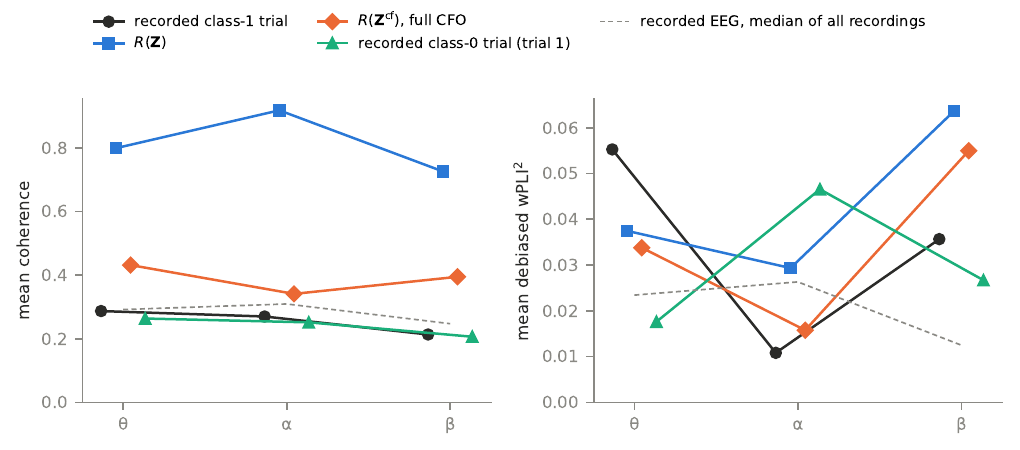}
    \caption{Mean coherence (left) and debiased wPLI$^2$ (right) over the 91 electrode pairs, by band, for the arousal case recording (participant 19, trial 2): the recorded high-arousal trial, its unedited reconstruction $R(\mathbf Z)$, the full-CFO counterfactual $R(\mathbf Z^{\mathrm{cf}})$, and a recorded low-arousal trial of the same participant (trial 1). The dashed line is the median over all recorded arousal trials.}
    \label{fig:connectivity}
\end{figure}

The decoder also distorts connectivity between electrodes, and the physiological penalties correct most of this distortion. Across tasks and frequency bands, full CFO removed 60--79\% of the increase in median coherence introduced by reconstruction (Table~\ref{tab:connectivity}); coherence is averaged over the 91 electrode pairs within each trial before taking the median. For example, median alpha coherence in the arousal task rose from 0.309 in recorded EEG to 0.946 after reconstruction. Full CFO brought it down to 0.466, closing 75\% of that gap, and untargeted repair brought it down to 0.479, closing 73\%. Conditions without the physiological penalties stayed at the values of the unedited reconstruction. Coherence after full CFO (0.37--0.52 across bands) remains above that of recorded EEG (0.25--0.34). Debiased wPLI$^2$ also moved toward the recorded values under full CFO, reaching 0.021--0.031 across tasks and bands, close to the recorded range in theta and alpha and about twice the recorded level in beta. The arousal case showed the same pattern (Figure~\ref{fig:connectivity}). Across bands, coherence fell from 0.73--0.92 in the unedited reconstruction to 0.34--0.43 after full CFO, against 0.21--0.29 and 0.21--0.26 for the participant's recorded high- and low-arousal trials. Together with the comparable reductions under untargeted repair, this indicates that the change corrects excess coherence introduced during reconstruction rather than shifting connectivity toward the target class.

\section{Conclusion}
We used the class distributions learned by a variational EEG classifier to generate counterfactual waveforms for valence and arousal. Across 23 held-out DREAMER participants, among correctly classified trials whose original prediction survived unedited reconstruction, the requested class change after decoding and re-encoding occurred in 82.0\% of valence trials and 56.7\% of arousal trials with full CFO, compared with 26.3\% and 9.1\% with base CFO. Base CFO used at most 50 updates and stopped at the first candidate whose latent target-class probability was at least 0.65 and whose decoded waveform passed the enabled physiological checks. Full CFO added Gaussian typicality regularization, could continue optimizing after reaching feasibility, and selected the feasible candidate with the lowest Gaussian discrepancy. The observed gains reflect these changes jointly. 

The resulting waveforms also more often reached a central region of the modeled target class. We defined this region using the 10th percentile of Gaussian discrepancy among real target-class recordings from the training participants. The fraction of eligible trials meeting this criterion rose from 5.1\% with base CFO to 58.4\% with full CFO for valence, and from 4.1\% to 40.1\% for arousal. With the full objective, 33.6\% of valence trials and 19.4\% of arousal trials met both this criterion and a re-encoded target probability of at least 0.65. Removing the physiological penalties raised the fraction meeting both criteria above 99\% for each task, while reducing agreement with the physiological reference ranges. On the same recordings, CoMTE and full CFO reached the target class at similar rates. Compared with full CFO, CoMTE's mean physiology grade was slightly lower, while its mean waveform RMSE was 2.8--2.9 times as high relative to the unedited reconstruction and 17--18\% lower relative to the recorded EEG. These results show that a decoded EEG waveform can illustrate what the classifier treats as central to an affective class, and that confidence and centrality alone provide an incomplete account of physiological plausibility. The counterfactuals also revealed a sensitivity to EEG window edges that whole-trial band summaries can obscure. Retaining only the edits in the first and last 250 ms of each six-second window preserved all 202 tested prediction flips across valence and arousal. Independent validation of the generated patterns against measured changes in affect is therefore an important next step before using them to guide neurofeedback.

\newpage

\section*{Acknowledgments}

This research was supported in part by EMOTIV\ and the UNC Graduate Fund. The authors would also like to thank Kerem Zaman for the insightful discussions that contributed to the development of this study.

\section*{Reproducibility}

Code developed for this study contributed substantially to the open-source \textit{EEGProc} library (\url{https://github.com/EEGo-UNC/EEGProc}). The model implementations and experiment scripts used in this study will be released in the \textit{EEGProc-XAI-CFO} repository (\url{https://github.com/EEGo-UNC/EEGProc-XAI-CFO}). The repository's Markdown documentation will provide instructions for running the experiments with the hyperparameter settings used in this study.

\section*{AI Use Statement}

Generative AI tools (ChatGPT6-Astra, ChatGPT5.6-Sol, Claude Fable 5.1) were used to assist with manuscript review, literature discovery, and the implementation of this study. This included assistance with code generation and debugging. The authors reviewed and refined all AI-assisted contributions.

\bibliography{2027_conf}
\bibliographystyle{2027_conf}

\newpage

\appendix
\section{Deep Learning Model}
\subsection{Adaptation of Variational Classification Objective}
\label{app:adapted-vc}

We adapt the variational classification objective of
\citet{dhuliawala2023variational} by representing the empirical and
anticipated class-conditional latent distributions as diagonal Gaussians:
\begin{equation}
q_{\phi}(z\mid y)
=
\mathcal N\!\left(m_y,\operatorname{diag}(v_y)\right),
\qquad
p_{\theta}(z\mid y)
=
\mathcal N\!\left(\mu_y,\operatorname{diag}(\sigma_y^2)\right).
\end{equation}

For each six-second training window $i$, let $Z_i$ denote its encoded sequence
and let $z_i=C_\theta(Z_i)\in\mathbb R^d$ denote the terminal
embedding supplied to the variational classifier. The distributions
$q_\phi(z\mid y)$ and $p_\theta(z\mid y)$ are defined in the corresponding
classification-embedding space. For each class label
$y\in\mathcal Y=\{0,1\}$, let $B_y$ denote the indices of observations
with label $y$ in the training minibatch. The coordinate-wise
empirical Gaussian parameters are, for $j=1,\ldots,d$,
\begin{equation}
m_{y,j}
=
\frac{1}{|B_y|}
\sum_{i\in B_y} z_{i,j},
\qquad
v_{y,j}
=
\frac{1}{|B_y|}
\sum_{i\in B_y}
\left(z_{i,j}-m_{y,j}\right)^2.
\end{equation}

This replaces discriminatory posterior density estimation with an analytic Gaussian KL divergence. Using a dimension-normalized posterior term, the resulting
objective is
\begin{equation}
\begin{aligned}
\mathcal L_{\mathrm{VC}}
={}&
\frac{\beta}{2d}
\sum_{y\in\mathcal Y}p(y)
\sum_{j=1}^{d}
\left[
\log\frac{\sigma_{y,j}^{2}}{v_{y,j}}
+
\frac{
v_{y,j}+(m_{y,j}-\mu_{y,j})^{2}
}{
\sigma_{y,j}^{2}
}
-1
\right]
+
\lambda
\sum_{y\in\mathcal Y}
p(y)\log\frac{p(y)}{p_{\pi}(y)}.
\end{aligned}
\label{eq:adapted_vc_closed_form}
\end{equation}
Here, $d$ is the latent dimensionality, $p(y)$ is the empirical frequency of label $y$, and $p_{\pi}(y)$ is its learned categorical prior. The $-\mathcal{H}(q)$ contribution in Equation~\ref{eq:vc_reg} accounts for the negative entropy of the posterior, while the remaining Gaussian terms align its mean and variance with the learned class-conditional distribution. In practice, all variances are lower-bounded by a small constant $\varepsilon>0$ for numerical stability.

\subsection{Inverse Subject-Adversarial}
We encourage cross-subject generalization through a subject-adversarial objective inspired by \citet{ozdenizci2020learning}. For window $w$ of trial $t$ from participant $s$, the adversary receives the classification embedding $u_{s,t,w}=C_\theta(\mathbf Z_{s,t,w})$, formed from the BiGRU's final forward and backward states. It does not receive a separately pooled trial representation. A dense subject head maps this embedding to a softmax over the 22 source identities in each LOSOCV fold. The task and adversarial objectives are

\begin{equation}
(\hat{\theta}_e,\hat{\theta}_c,\hat{\theta}_a)
=
\arg\min_{\theta_e,\theta_c}\max_{\theta_a}
\left\{
-\mathbb{E}_{(X,y)}\!\left[\log q_{\theta_c}(y\mid u)\right]
+\lambda\mathbb{E}_{(X,s)}\!\left[\log q_{\theta_a}(s\mid u)\right]
\right\},
\label{eq:adv_objective}
\end{equation}

where $u=C_\theta(g_\phi(X))$, $\theta_e$ denotes the upstream parameters producing this embedding, $\theta_c$ denotes the affective prediction head's parameters, $\theta_a$ denotes the subject head's parameters, and $\lambda>0$ controls the adversarial contribution.

We implement this objective with a gradient reversal layer \citep{ganin2016domain}. The subject head minimizes identity cross-entropy, while the encoder and recurrent summarizer receive its negatively scaled gradient alongside the other joint gradients. The objective discourages identity information recoverable from the window classification embedding; it does not establish independence of the complete latent sequence from participant identity. We average the window probabilities for emotion-recognition evaluation and counterfactual analysis, as specified in Equation~\ref{eq:trial_probability_mean}.

\subsection{Supervised Contrastive Learning}
\label{app:sup}

To encourage emotion-related representations that are shared across subjects during autoencoding, we adapt the supervised contrastive objective of \citet{khosla2020supervised}. The objective uses emotion labels to identify positive pairs and subject identities to restrict these pairs to observations
from different participants. For a mini-batch indexed by $\mathcal{M}=\{1,\ldots,B\}$, let $z_i$ denote the learned representation of sample $i$, with emotion label $y_i$ and subject identity $s_i$. We normalize the representations as $u_i=z_i/\lVert z_i\rVert_2$, so that $u_i^\top u_j$ measures cosine similarity. For each anchor $i$, the positive and eligible comparison sets are
\begin{equation}
\begin{aligned}
P(i) &= \{p\in\mathcal{M}\setminus\{i\}: y_p=y_i,\ s_p\neq s_i\},\\
J(i) &= P(i)\cup\{j\in\mathcal{M}: y_j\neq y_i\}.
\end{aligned}
\label{eq:sup_pair_sets}
\end{equation}
Thus, same-label samples from the same subject are excluded from both the
positive set and the denominator, while different-label samples remain
eligible negatives regardless of subject identity. The anchor itself is
always excluded.

Writing $\mathcal{M}_{+}=\{i\in\mathcal{M}: |P(i)|>0\}$ for the valid
anchors, the supervised contrastive loss is
\begin{equation}
\mathcal{L}_{\mathrm{sup}}
=
-\frac{1}{\sum_{i\in\mathcal{M}_{+}} w_i}
\sum_{i\in\mathcal{M}_{+}}
\frac{w_i}{|P(i)|}
\sum_{p\in P(i)}
\log\!\left[
\frac{\exp(u_i^\top u_p/\tau)}
{\sum_{j\in J(i)}\exp(u_i^\top u_j/\tau)}
\right],
\label{eq:sup_objective}
\end{equation}
where $\tau>0$ is the temperature parameter and $w_i\geq0$ is an optional anchor weight. Setting $w_i=1$ recovers the unweighted mean over valid anchors. If no valid anchors exist, or their total
weight is zero, the loss is defined as zero.

Minimizing $\mathcal{L}_{\mathrm{sup}}$ encourages same-emotion samples from different participants to have similar representations relative to samples with different emotion labels. Our adaptation differs from the standard supervised contrastive objective by conditioning positive-pair selection on subject identity and excluding same-subject, same-label comparisons entirely. This prevents repeated windows from a single
participant from supplying the positive-pair signal and avoids treating those excluded windows as negatives. Because cross-subject alignment alone does not guarantee that all subject-specific information has been removed, this term is complementary to subject-adversarial learning.

\subsection{LOSO Emotion Recognition}\label{app:model_report}

The subject-wise results and plots below describe trial-level emotion recognition at epoch 50 across all 23 LOSOCV folds.

\subsubsection{Balanced Accuracy as the Primary Metric}\label{app:bacc}
Ordinary accuracy can overstate performance in imbalanced EEG emotion recognition tasks. The majority-class rates in the DREAMER dataset are 72.5\% for high arousal and 61.1\% for high valence. Many studies do not report balanced metrics. For example, MTLFuseNet reports LOSOCV accuracies of 80.43\% for valence and 83.33\% for arousal \citep{li2023mtlfusenet}. The paper reports no BA, and we could not locate a public implementation or reproduce the results from the published description. We therefore treat these values as reported results rather than a reproducible benchmark. Similarly, CMHFE-DAN reports 64.73\% valence and 79.83\% arousal accuracy, only 3.63\% and 7.33\% above the corresponding majority-class baselines \citep{hilali2025cmhfedan}. Because CMHFE-DAN also accesses unlabeled target-subject data during training, it represents transductive adaptation rather than strictly zero-shot evaluation \citep{hilali2025cmhfedan}. More broadly, performance is highly sensitive to preprocessing, potential artifact exploitation, and validation settings \citep{delpup2025morebetter}. Meanwhile, methodological problems such as leakage, biased feature selection, and model collapse were found in nearly 87\% of 101 reviewed studies and inflated accuracy by up to 46 percentage points \citep{khan2025reviewprocessfailures}. Under zero-shot LOSOCV on DREAMER, an RGB-style EEG model achieved 75.15\% accuracy but reported 58.73\% BA for arousal, and 65.55\% versus 60.05\% for valence \citep{zhang2026rgbstyle}. This large gap is consistent with a strong majority-class bias. The corresponding valence metrics, while less pronounced, also suggest the possibility of class imbalance. Because the reported LOSOCV metrics are averaged across subjects and class-specific recalls are not provided, the extent of minority-class under-recognition cannot be determined exactly. We use trial-level BA as the main metric and report class recalls separately. We report BA at a fixed threshold and at participant-specific oracle thresholds selected on the held-out trials.

\subsubsection{Predictive Model Evaluation Metrics}

We compute confusion counts over trials to assess recognition of both high and low ratings for valence and arousal. Using these counts, we define the true positive rate ($\mathrm{TPR}$) and the false positive rate ($\mathrm{FPR}$) as

\begin{equation}
\begin{aligned}
\mathrm{TPR} &= \frac{TP}{TP + FN},
&\quad
\mathrm{FPR} = \frac{FP}{FP + TN}. \\
\end{aligned}
\end{equation}

Then we evaluate the recall of each class:

\begin{equation}
\begin{aligned}
\mathrm{Recall}_{0} &= 1 - \mathrm{FPR}, \qquad
\mathrm{Recall}_{1} = \mathrm{TPR}.
\end{aligned}
\end{equation}

Finally, we evaluate BA:

\begin{equation}
\begin{aligned}
\mathrm{BA} &= \frac{1}{2}
\left(\mathrm{Recall}_{0} + \mathrm{Recall}_{1}\right),
\end{aligned}
\end{equation}

\subsubsection{Discussion of Results}
\label{app:trial_subject_results}

\begin{figure}[h]
    \centering
    \includegraphics[width=0.48\columnwidth]{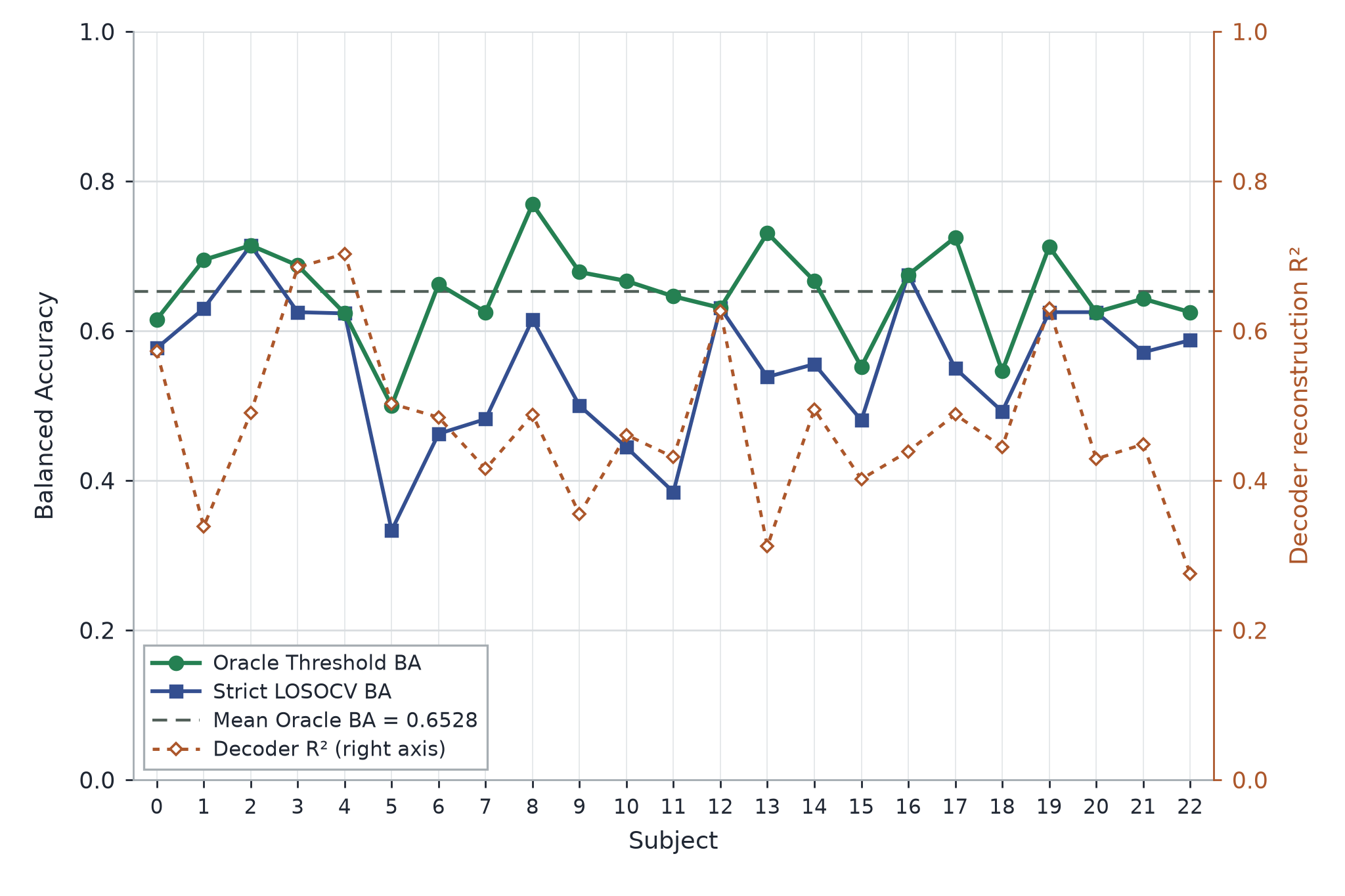}
    \hfill
    \includegraphics[width=0.48\columnwidth]{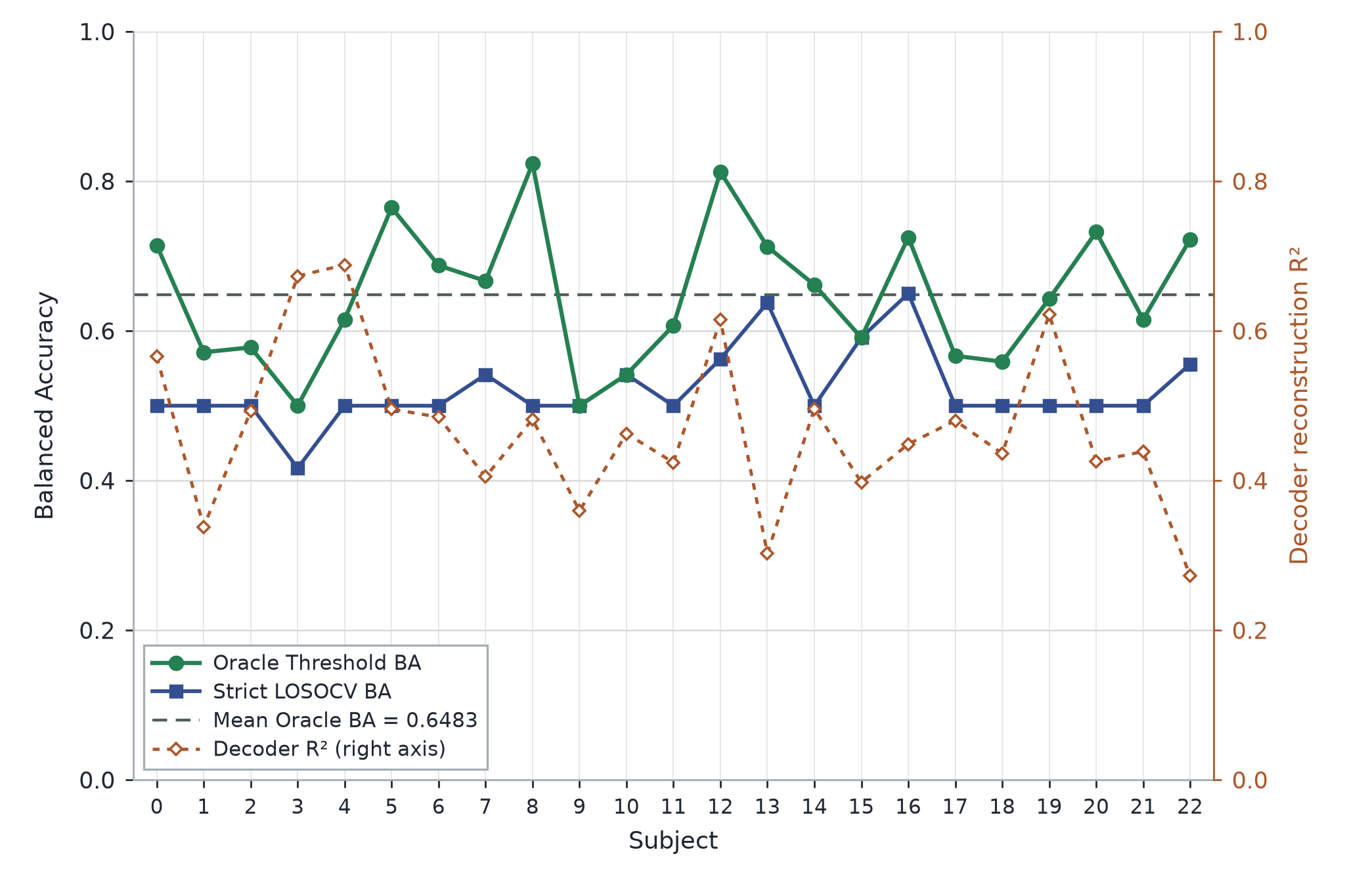}
    \caption{Trial-level balanced accuracy for valence (left) and arousal (right). Green circles show results at participant-specific oracle thresholds, and blue squares show strict LOSOCV at a fixed cutoff. The horizontal dashed line marks mean oracle balanced accuracy. Orange diamonds show decoder reconstruction $R^2$ on the right axis. Each participant contributed 18 trials.}
    \label{fig:losocv_folds}
\end{figure}

Figure~\ref{fig:losocv_folds} shows trial-level BA at participant-specific oracle thresholds and under strict LOSOCV with a fixed cutoff. The right axis shows held-out decoder reconstruction $R^2$. At epoch~50, we averaged the ten window probabilities within each trial and selected a separate threshold for each participant and task by maximizing BA on their 18 labeled trials. Mean oracle BA was 65.28\% for valence and 64.83\% for arousal. Table~\ref{tab:emotion_recognition} reports a separate fixed-threshold analysis using one cutoff per task selected on the pooled out-of-fold predictions. Individual oracle BA ranged from 50.00\% (subject~5) to 76.92\% (subject~8) for valence and from 50.00\% (subjects~3 and~9) to 82.35\% (subject~8) for arousal.

The number of low-rated trials ranged from 4 to 11 per participant for valence and from 1 to 9 for arousal. Subjects~5, 8, and~18 each had one low-arousal trial and 17 high-arousal trials. At their oracle thresholds, subjects~8 and~18 both correctly classified their single low-arousal trial. Subject~8 classified 11 of 17 high-arousal trials correctly (64.71\% recall), and subject~18 classified 2 of 17 correctly (11.76\% recall). Their resulting balanced accuracies were 82.35\% and 55.88\%, and their overall accuracies were 66.67\% and 16.67\%, respectively.

For a class represented by $n_c$ trials, changing one trial prediction from incorrect to correct at a given threshold increases BA by $1/(2n_c)$. For the three participants with one low-arousal trial, that trial determines their entire low-arousal recall and can therefore change BA by 50 percentage points. Their oracle scores are sensitive to the prediction on that single trial.

Decoder $R^2$ measures EEG reconstruction fidelity. Mean held-out $R^2$ was 0.4747 for valence and 0.4698 for arousal (Table~\ref{tab:emotion_recognition}). Joint decoder training adds a waveform reconstruction objective to the shared encoder. Preserving waveform detail may compete with learning features that distinguish emotion classes, making emotion recognition more difficult.

\section{Counterfactual Optimization}
\subsection{Non-increasing Subject Identifiability in Counterfactual Optimization}
\label{app:proof_of_sub_inv}

Let $s\in\mathcal S$ denote participant identity and $\mathbf Z=(\mathbf Z_1,\ldots,\mathbf Z_W)$ the complete trial's latent windows. ISA predicts identity separately from each window classification embedding $C_\theta(\mathbf Z_w)$, with loss $\ell_a(\mathbf Z_w,s)=-\log q_a(s\mid C_\theta(\mathbf Z_w))$. The full-trial affective prediction averages window probabilities; the identity head does not operate on a pooled trial representation. Gradient reversal encourages reduced identity recovery from these embeddings but does not guarantee independence between $\mathbf Z$ and $s$.

For a fixed trained model, reference quantities, and optimizer randomness, collect the trial-specific inputs to CFO and physical-unit decoding as $U=(\mathbf Z,\mathbf r)$, where $\mathbf r$ contains the retained per-window RMS scales. Assume all trial-dependent target selection, objective evaluation, stopping, and candidate selection use only $U$. Then $\mathbf Z^{\mathrm{cf}}=T(U)$ and the physical-unit counterfactual is $\mathbf X^{\mathrm{cf}}=G(U)=\mathbf r\odot R(T(U))$.

Define the optimal subject-identification accuracy of a representation $U$ as
\begin{equation}
\operatorname{Acc}^{*}_{s}(U)=\sup_b\Pr[b(U)=s],
\end{equation}
where the supremum is over all measurable subject predictors. Any predictor $b$ applied to $T(U)$ induces a predictor $b\circ T$ applied to $U$. Hence,
\begin{equation}
\begin{aligned}
\operatorname{Acc}^{*}_{s}(\mathbf Z^{\mathrm{cf}})
&=\sup_b\Pr[b(T(U))=s]\leq\sup_c\Pr[c(U)=s]
=\operatorname{Acc}^{*}_{s}(U).
\end{aligned}
\label{eq:subinv_proof}
\end{equation}
The same composition argument applied to the complete mapping $G$ gives
\begin{equation}
\operatorname{Acc}^{*}_{s}(\mathbf X^{\mathrm{cf}})
\leq\operatorname{Acc}^{*}_{s}(U)
=\operatorname{Acc}^{*}_{s}(\mathbf Z,\mathbf r).
\end{equation}
These comparisons use the same population of trials under the same fixed model. The bound concerns the information available in the complete input $U$ to counterfactual generation and physical-unit reconstruction. It does not establish that ISA achieved low optimal subject identifiability, or that a low score from the trained window-level identity head characterizes all identity information in $U$.

\subsection{Construction of Volume-Conduction Spatial Coherence}
\label{app:vcsc}

We estimate connectivity from the ten six-second windows of each trial.
Let $X^w_{i,b}(f)$ denote the discrete Fourier transform (DFT) of the
band-$b$ signal at electrode $i$ in window $w$. For electrode pair $(i,j)$,
define the cross-spectrum and its imaginary part as
\[
S^w_{ij,b}(f)=X^w_{i,b}(f)\overline{X^w_{j,b}(f)}, \qquad
I^w_{ij,b}(f)=\operatorname{Im} S^w_{ij,b}(f).
\]
At each frequency, we treat the windows as repeated spectral observations.
Writing $\langle\cdot\rangle_w$ for the average over windows and
$\langle\cdot\rangle_{f\in b}$ for the average over in-band DFT bins,
we compute magnitude-squared coherence and debiased squared wPLI
\citep{vinck2011improved} as
\begin{equation}
\begin{aligned}
C^b_{ij}
&=\left\langle
\frac{|\langle S^w_{ij,b}(f)\rangle_w|^2}
     {\langle |X^w_{i,b}(f)|^2\rangle_w\,
      \langle |X^w_{j,b}(f)|^2\rangle_w}
\right\rangle_{f\in b},\\
\Omega^b_{ij}
&=\left\langle
\frac{\bigl(\sum_w I^w_{ij,b}(f)\bigr)^2-\sum_w I^w_{ij,b}(f)^2}
     {\bigl(\sum_w |I^w_{ij,b}(f)|\bigr)^2-\sum_w I^w_{ij,b}(f)^2}
\right\rangle_{f\in b}.
\end{aligned}
\end{equation}
For $I_w=I^w_{ij,b}(f)$, the wPLI denominator equals
$\sum_{w\ne v}|I_w|\,|I_v|\geq 0$. The coherence denominator is also
nonnegative. The implementation adds a small positive constant to each
denominator for numerical stability.

VCSC uses untapered DFTs of RMS-normalized windows. We fit its reference
matrices on real source trials using the same segmentation and normalization.
Let $A_{ij}$ be the combined standardized deviation for pair $(i,j)$.
For a decoded candidate $R(\mathbf Z')$, the VCSC penalty is
\begin{equation}
\begin{aligned}
A_{ij}
&=\left\{\sum_b\left[
\left(\frac{C^b_{ij}-\widehat C^b_{ij}}{\sigma^b_{C,ij}}\right)^2+
\left(\frac{\Omega^b_{ij}-\widehat\Omega^b_{ij}}{\sigma^b_{\Omega,ij}}\right)^2
\right]\right\}^{1/2},\\
\mathcal P_{\mathrm{ph}}\!\left(R(\mathbf Z')\right)
&=\frac{1}{|\mathcal E|}\sum_{(i,j)\in\mathcal E}
w(d_{ij})\,\phi(A_{ij}).
\end{aligned}
\label{eq:VCSC}
\end{equation}
Here $\mathcal E$ contains the 91 electrode pairs and $d_{ij}$ is the
three-dimensional chord distance between their positions. With
$x^+=\max(x,0)$, we use $w(d)=\exp((d_c-d)^+/\tau)$, where
$d_c=12\,\mathrm{cm}$ and $\tau=4\,\mathrm{cm}$. Pairs closer than $d_c$
receive greater weight. The penalty function is
\begin{equation}
\phi(a)=\exp\!\bigl((\min(a,A_{\max})-A_0)^+\bigr)-1
       +\exp\!\bigl((A_{\max}-A_0)^+\bigr)(a-A_{\max})^+ .
\end{equation}
With $A_0=2$ and $A_{\max}=20$, the penalty is zero up to $A_0$, grows
exponentially up to $A_{\max}$, and continues linearly with the same slope.
We calibrate the VCSC feasibility tolerance from real source-trial VCSC
scores. For the componentwise connectivity assessment below, we use
demeaned, Hann-tapered windows in physical units and fit separate source
reference intervals.

\subsection{Graded Physiological Assessment}
\label{app:graded_physiology}

\paragraph{Amplitude and band-power penalty.}
We restore physical units using the original per-window RMS scales,
$\mathbf X'=\mathbf r\odot R(\mathbf Z')$. For each electrode--band
combination, amplitude is the 99th percentile of absolute values across
the complete trial. To estimate band power, we detrend each window, compute
its Hann-tapered periodogram, average the spectral densities across windows,
and integrate over the passband.

For each component $k$, let $[\ell_k,u_k]$ be its central 95\% interval
across real trials from the training participants in the fold. For intervals
with positive half-width, define
\[
m_k=\frac{\ell_k+u_k}{2},\qquad h_k=\frac{u_k-\ell_k}{2},\qquad
e_k=\max\!\left(0,\frac{|v_k-m_k|}{h_k}-1\right),
\]
where $v_k$ is the trial's component value. A value inside the interval
has $e_k=0$. A value one tenth of a half-width beyond the nearer edge
has $e_k=0.1$. Exceedance has no upper bound. For the set
$\mathcal K_{\mathrm{AP}}$ of defined amplitude and band-power components
with positive reference half-width, the optimization penalty is
\begin{equation}
\mathcal P_{\mathrm{AP}}\!\left(\mathbf r\odot R(\mathbf Z')\right)
=\frac{1}{|\mathcal K_{\mathrm{AP}}|}
 \sum_{k\in\mathcal K_{\mathrm{AP}}}e_k^2 .
\label{eq:amplitude_power_penalty}
\end{equation}

\paragraph{Trial grade and binary acceptance.}
We compute the same component exceedance for amplitude, band power,
magnitude-squared coherence, and debiased squared wPLI.
For family $f$, let $\mathcal K_f$ contain its scorable components.
We calculate the trial grade from the largest family mean exceedance,
\[
\bar e_f=\frac{1}{|\mathcal K_f|}\sum_{k\in\mathcal K_f}e_k,\qquad
\bar e_{\max}=\max_{f:\,|\mathcal K_f|>0}\bar e_f,\qquad
G=\frac{1}{1+\bar e_{\max}} .
\]
A grade of 1 indicates that every scorable component lies within its
reference interval. We report the mean of the individual trial grades
in Table~\ref{tab:counterfactual results}.

For binary assessment, we count the fraction of a family's defined components
that lie within their reference intervals. Because the intervals come from other
participants, recorded EEG of a held-out participant often falls slightly outside a few
of them: under a flat requirement of 95\% of components, only 64--69\% of the recorded
held-out trials pass both the amplitude and band-power checks. For each outer LOSOCV fold, we used only its 22 training participants to construct
the physiological reference intervals and calibrate the amplitude and band-power
pass thresholds. For each family, the threshold was the 5th percentile of the
within-interval fraction across the training participants' recorded trials.
We applied the resulting thresholds to the held-out participant's recordings
and counterfactuals. The required fraction
is 0.71--0.79 for amplitude and 0.67--0.76 for band power, and 93--94\% of the recorded
trials pass both checks. Defined zero-width intervals
enter this binary assessment and are excluded from the graded calculation.
Only defined components can contribute a pass. Joint physiological
acceptance requires all four family checks and the calibrated VCSC
feasibility check. The band-filtered decoder does not support estimation
of the aperiodic exponent.

\subsection{Architecture-Agnostic Implementation}
We provide a reference implementation of the counterfactual optimizer through the \textit{EEGProc} Python library. The implementation separates the optimization procedure from model-specific architecture through a common adapter interface. For a given trained model, the adapter exposes the operations required to (i) obtain the latent representation, (ii) evaluate target-class probabilities from a candidate latent sequence, and (iii) decode that sequence into the input feature domain. Model-specific adapters remain responsible for tensor-shape conventions, preprocessing and inverse transformations. Thus, the implementation is architecture-agnostic subject to the differentiable latent-classification-decoding interface assumed by our method. It is not intended to universally support black-box predictors.

\subsection{Counterfactual Experimental Details}
\label{app:cfo_experimental_details}

The experiments use reconstruction-trained window models to examine decoded validity, physiological repair, support from real target-class EEG, and recoverable subject information through a counterfactual suite and companion analyses. The fixed six-second model configuration is shared across all LOSOCV folds, and the counterfactual suite uses the frozen epoch-50 checkpoints.

\subsubsection{Full-Trial Counterfactual Suite}
\label{app:cfo_window_suite}

For each task, the suite uses the frozen LOSOCV model for the corresponding held-out participant. Each central 60-s recording supplies 10 non-overlapping six-second windows, each with 768 samples and 42 electrode--band features. A complete trial is eligible when its mean window probability yields the correct recording label. Each eligible trial targets the opposite class. The suite records both directions separately, and the reported comparisons pool directions within each task. The checkpoint, preprocessing, and eligibility set are shared across all conditions, and the complete model remains frozen. Physical-unit physiological measurements restore each original window's RMS scale, while VCSC uses the matched normalized full-trial estimator.

The objectives in Equations~\ref{eq:counterfactual-objective} and~\ref{eq:typicality_cfo_objective} jointly optimize all ten windows against the trial-averaged target. The physiological component combines source-calibrated VCSC and amplitude/band-power penalties. The six conditions are:
\begin{enumerate}
    \item \textbf{Target and latent-proximity baseline ($\mathcal{L}_{\mathrm{base}}$).} Only the target loss and latent-proximity penalty are active:
    \begin{equation}
    \mathcal{L}_{\mathrm{base}}
    =\lambda_o\mathcal{L}_{\mathrm{o}}\bigl(f(\mathbf{Z'}),y_{\mathrm{o}}\bigr)
    +\lambda_zd_z(\mathbf{Z},\mathbf{Z'}).
    \label{eq:cfo_baseline}
    \end{equation}
    Decoded proximity, physiology, and typicality have zero weight.

    \item \textbf{CFO without typicality ($\mathcal{L}_{\mathrm{cfo}}$).} Target, latent-proximity, decoded-proximity, VCSC, and amplitude/band-power terms are active. Compared with the latent-only baseline, this adds decoded proximity and physiology together.

    \item \textbf{Full CFO with typicality ($\mathcal{L}_{\mathrm{cfo}}^{\mathcal{T}_{C_c}}$).} All terms are active, including normalized discrepancy from the requested target-class Gaussian. The comparison with $\mathcal{L}_{\mathrm{cfo}}$ includes the discrepancy penalty, continued refinement after feasibility, and endpoint selection by discrepancy among feasible candidates. Base CFO stops at first feasibility. Mean completed updates for base and full CFO were 45.6 and 49.0 for valence and 46.7 and 48.9 for arousal.

    \item \textbf{Typicality without physiology ($\mathcal{L}_{\mathrm{cfo}}^{\mathcal{T}_{C_c}-\lambda_{\mathrm{ph}}}$).} The full objective's VCSC and amplitude/band-power weights are set to zero, retaining target, both proximity terms, and typicality. Disabling VCSC also removes its requirement for activating typicality, so this ablation changes both the penalty and the activation rule.

    \item \textbf{Decoded-proximity-only ablation ($\mathcal{L}_{\mathrm{cfo}}^{\mathcal{T}_{C_c}-\lambda_z}$).} The full objective's latent-proximity weight is set to zero, while decoded proximity remains active with weight $\lambda_R=0.1$. Target, VCSC, amplitude/band-power, and typicality terms are retained. Decoded proximity is therefore the only proximity constraint. Comparison with the full objective tests the latent-proximity term's contribution.

    \item \textbf{Physiology-only repair ($\mathcal{L}_{\mathrm{repair}}$).} Target and typicality weights are zero, while latent proximity, decoded proximity, VCSC, and amplitude/band-power regularization remain active. This control tests how much decoded change also occurs without a class-targeting objective. Its endpoints are evaluated for physiological feasibility and preservation of the original prediction, rather than class-flip success.
\end{enumerate}

Unless disabled by the specified ablation, active weights are $1$ for the target, $0.1$ for each proximity term, $1$ for each of VCSC and amplitude/band power, and $1$ for typicality. All conditions share initialization, per-trial seeds, learning rate $0.01$, and a maximum of 50 optimization updates. The target-confidence threshold is $0.65$. Optimization feasibility requires latent trial target assignment and confidence together with the enabled VCSC and amplitude/power checks; repair omits the target requirement. Decoded confidence and the independent componentwise connectivity checks are endpoint evaluations, not stopping gates. The baseline, CFO-without-typicality, and repair conditions stop at first optimization feasibility. Typicality conditions activate after latent trial target confidence and enabled VCSC feasibility, then continue to refine discrepancy; amplitude/power remains part of the loss and final feasibility. Searches also stop at the update budget or gradient stagnation, and typicality refinement may stop after a feasible discrepancy plateau. Thus, maximum budgets match while actual iteration counts can differ. Every class-targeting search starts from the original latent representation, not the repair endpoint. Alongside the primary confidence criterion, the suite records target-probability passage at thresholds $0.50$, $0.55$, $0.60$, and $0.65$ in both class directions, separately in latent space and after decoding and re-encoding.

For each requested class $c$, the Gaussian discrepancy in Equation~\ref{eq:class1_latent_discrepancy} averages scores across all ten windows and uses the source target-class trial 10th-percentile cutoff. Generated waveforms are decoded, re-encoded, and classified to measure decoded target assignment, confidence, and Gaussian acceptance separately from latent success. We compare original EEG, reconstructions, counterfactuals, and repair endpoints, including reconstruction-to-repair and repair-to-counterfactual RMSE.

\subsubsection{Comparison with CoMTE}
\label{app:cfo_comte}

We compare adapted CoMTE \citep{ates2021comte} with CFO using the same frozen window classifier, mean trial probabilities, preprocessing, eligible complete trials, targets, confidence thresholds, and endpoint evaluation. Each candidate replaces all three band waveforms of selected electrodes across the full trial with the corresponding waveforms of one donor. The two nearest correctly classified target-class donor trials come only from source participants. Search uses a soft three-electrode limit, five additional hill-climb restarts, and at most 2,000 full-trial classifier queries per attempt. CFO permits at most 50 gradient updates. These budgets count different operations: a gradient update includes backward computation, so query/update counts alone do not establish a runtime advantage. CoMTE averaged 1,992.7 classifier queries per valence trial and 1,995.8 per arousal trial. Full CFO averaged 49.0 and 48.9 gradient updates, respectively. Mean generation times per trial were 132.9~s for CoMTE and 106.5~s for full CFO on valence, and 187.4~s and 108.9~s, respectively, on arousal. CoMTE times its donor search, ending before endpoint evaluation. The CFO timer includes input encoding, optimization and optimizer-side endpoint diagnostics. These figures describe the recorded runs.

CoMTE has direct access to observed target-class donor waveforms during generation, requires only prediction queries, and avoids a decoder and its reconstruction error. Electrode-group substitution also makes the changed channels explicit and preserves the other channels exactly. These properties may favor amplitude/band-power scores and signal fidelity compared with decoded CFO. They may also provide low displacement scores that are, nevertheless, unwanted due to selective substitution of small feature subsets. They do not guarantee physiological passage as donor values need not lie inside every source interval, and combining electrodes from different recordings can disrupt cross-electrode coherence or phase relationships.

Conversely, this CoMTE adaptation searches a discrete set of electrode substitutions from only two candidate donors and cannot synthesize intermediate temporal or spectral edits. Its search objective targets confidence and a soft electrode-count penalty, without CFO's explicit class-Gaussian, latent-proximity, or physiological regularization. CFO offers continuous latent changes and those regularizers, but requires differentiable model access and a trained decoder; reconstruction can distort the delivered signal and latent edits need not be sparse in electrode space. Their computational costs are reported under the stated method-specific budgets, and failed attempts remain in the denominators.

\subsection{Paired Optimized Discrepancy}\label{app:discrepancy}

For each eligible complete trial $i$ targeting class $c\in\{0,1\}$, let $\mathbf Z_i^{\mathrm{cf},(a)}$ denote the latent endpoint under class-targeting objective $a$. Let $D_{C_c}$ be the target-class discrepancy and $\tau_c>0$ the 10th percentile of source target-class discrepancies for the frozen model. We compare an alternative objective $\mathrm{cfo}_m$ with the CFO objective in Equation~\ref{eq:counterfactual-objective} on the same complete trial, target class, and checkpoint:
\begin{equation}
\frac{\Delta D_{i,c}}{\tau_c}
=
\frac{
D_{C_c}\!\left(\mathbf Z_i^{\mathrm{cf},(\mathrm{cfo}_m)}\right)
-
D_{C_c}\!\left(\mathbf Z_i^{\mathrm{cf},(\mathrm{cfo})}\right)
}{\tau_c}.
\end{equation}
A negative value indicates lower target-class discrepancy under the alternative objective. This comparison uses latent endpoints; Gaussian acceptance of decoded signals is evaluated separately after re-encoding (Section~\ref{sec:distro_typ}).

\subsection{Counterfactual Results}\label{app:cfo_results}


\begin{table}[H]
\centering
\caption{Decoded target-class validity, waveform displacement, and graded physiology,
pooled across both target directions. Each validity entry gives successful trials divided by
eligible trials (percentage). Initially stable trials retained their original
predicted class after unedited reconstruction. For the untargeted repair
control, target hits are unintended class changes. Mean RMSE compares the
output EEG with the unedited reconstruction in normalized model units.
Both methods cover all 23 participants and the same 214 valence and 217 arousal trials. Section~\ref{sec:cfo_results_draft} also reports RMSE relative to the recorded EEG for full CFO and CoMTE. The physiology grade is $1/(1+\bar e_{\max})$,
where $\bar e_{\max}$ is the largest mean exceedance across scorable physiological
families in source half-interval widths. We report the mean grade across trials.}
\vspace{1mm}
\label{tab:counterfactual results}
\small
\setlength{\tabcolsep}{4pt}
\resizebox{\linewidth}{!}{%
\begin{tabular}{@{}lrrrrr@{}}
\toprule
Condition & \shortstack{Re-encoded\\validity}
          & \shortstack{Target hits on\\initially stable trials}
          & \shortstack{Target confidence\\$\geq 0.65$}
          & \shortstack{Mean\\RMSE} & \shortstack{Mean physiology\\grade (0--1)} \\
\midrule
\multicolumn{6}{@{}l}{\textit{Valence}} \\
Target + latent proximity & 84/214 (39.3\%) & 3/133 (2.3\%) & 15/214 (7\%) & 0.002 & 0.847 \\
Base & 104/214 (48.6\%) & 35/133 (26.3\%) & 14/214 (6.5\%) & 0.271 & 0.961 \\
Full typicality & 190/214 (88.8\%) & 109/133 (82\%) & 101/214 (47.2\%) & 0.283 & 0.964 \\
Without latent proximity & 190/214 (88.8\%) & 109/133 (82\%) & 101/214 (47.2\%) & 0.283 & 0.964 \\
Without physiology & 214/214 (100\%) & 133/133 (100\%) & 214/214 (100\%) & 0.098 & 0.847 \\
Repair control & 92/214 (43\%) & 29/133 (21.8\%) & 15/214 (7\%) & 0.259 & 0.951 \\
CoMTE & 188/214 (87.9\%) & 114/133 (85.7\%) & 67/214 (31.3\%) & 0.791 & 0.930 \\
\midrule
\multicolumn{6}{@{}l}{\textit{Arousal}} \\
Target + latent proximity & 55/217 (25.3\%) & 3/164 (1.8\%) & 6/217 (2.8\%) & 0.003 & 0.835 \\
Base & 53/217 (24.4\%) & 15/164 (9.1\%) & 8/217 (3.7\%) & 0.278 & 0.955 \\
Full typicality & 146/217 (67.3\%) & 93/164 (56.7\%) & 63/217 (29\%) & 0.287 & 0.957 \\
Without latent proximity & 146/217 (67.3\%) & 93/164 (56.7\%) & 63/217 (29\%) & 0.287 & 0.957 \\
Without physiology & 215/217 (99.1\%) & 162/164 (98.8\%) & 215/217 (99.1\%) & 0.101 & 0.835 \\
Repair control & 47/217 (21.7\%) & 10/164 (6.1\%) & 7/217 (3.2\%) & 0.270 & 0.947 \\
CoMTE & 145/217 (66.8\%) & 98/164 (59.8\%) & 33/217 (15.2\%) & 0.830 & 0.912 \\
\bottomrule
\end{tabular}}
\end{table}

\begin{table}[H]
\centering
\caption{Target-class Gaussian acceptance, pooled across both target
directions. Entries give passing trials divided by eligible trials
(percentage). P10 and P25 are the 10th and 25th percentiles of the
Gaussian discrepancy for source trials in the target class. A trial
passes when its output EEG has a re-encoded discrepancy at or below
the cutoff. Joint acceptance also requires a target-class probability
of at least 0.65 on the same trial. Both methods cover all 23 participants and the same 214 valence and 217 arousal trials.}
\vspace{1mm}
\label{tab:gaussian_acceptance}
\small
\setlength{\tabcolsep}{4pt}
\begin{tabular}{@{}lrrrr@{}}
\toprule
& \multicolumn{2}{c}{Gaussian acceptance}
& \multicolumn{2}{c}{Gaussian acceptance + confidence} \\
\cmidrule(lr){2-3}\cmidrule(l){4-5}
Condition & P10 & P25 & P10 & P25 \\
\midrule
\multicolumn{5}{@{}l}{\textit{Valence}} \\
Target + latent proximity & 10/214 (4.7\%) & 20/214 (9.3\%) & 1/214 (0.5\%) & 3/214 (1.4\%) \\
Base                      & 11/214 (5.1\%) & 27/214 (12.6\%) & 2/214 (0.9\%) & 7/214 (3.3\%) \\
Full typicality           & 125/214 (58.4\%) & 166/214 (77.6\%) & 72/214 (33.6\%) & 92/214 (43\%) \\
Without latent proximity  & 126/214 (58.9\%) & 166/214 (77.6\%) & 73/214 (34.1\%) & 92/214 (43\%) \\
Without physiology        & 213/214 (99.5\%) & 214/214 (100\%) & 213/214 (99.5\%) & 214/214 (100\%) \\
Repair control            & 10/214 (4.7\%) & 24/214 (11.2\%) & 2/214 (0.9\%) & 5/214 (2.3\%) \\
CoMTE                     & 20/214 (9.3\%) & 53/214 (24.8\%) & 11/214 (5.1\%) & 28/214 (13.1\%) \\
\midrule
\multicolumn{5}{@{}l}{\textit{Arousal}} \\
Target + latent proximity & 11/217 (5.1\%) & 31/217 (14.3\%) & 0/217 (0\%) & 1/217 (0.5\%) \\
Base                      & 9/217 (4.1\%) & 26/217 (12\%) & 0/217 (0\%) & 1/217 (0.5\%) \\
Full typicality           & 87/217 (40.1\%) & 134/217 (61.8\%) & 42/217 (19.4\%) & 56/217 (25.8\%) \\
Without latent proximity  & 87/217 (40.1\%) & 134/217 (61.8\%) & 42/217 (19.4\%) & 56/217 (25.8\%) \\
Without physiology        & 217/217 (100\%) & 217/217 (100\%) & 215/217 (99.1\%) & 215/217 (99.1\%) \\
Repair control            & 9/217 (4.1\%) & 25/217 (11.5\%) & 0/217 (0\%) & 1/217 (0.5\%) \\
CoMTE                     & 15/217 (6.9\%) & 54/217 (24.9\%) & 4/217 (1.8\%) & 10/217 (4.6\%) \\
\bottomrule
\end{tabular}
\end{table}

\begin{table}[H]
\centering
\caption{Prediction preservation and unwanted flips after unedited
reconstruction and physiological repair. A flip occurs when the re-encoded
EEG receives a different predicted class from the original EEG. The first
two rates use all eligible trials. The last two use only unwanted flips,
with $p$ denoting the probability assigned to the new class.}
\label{tab:prediction_preservation}
\small
\setlength{\tabcolsep}{4pt}
\resizebox{\linewidth}{!}{%
\begin{tabular}{@{}llrrrr@{}}
\toprule
& & \multicolumn{2}{c}{Among all trials}
  & \multicolumn{2}{c}{Among unwanted flips} \\
\cmidrule(lr){3-4}\cmidrule(l){5-6}
Task & Stage
     & \shortstack{Original prediction\\preserved}
     & Unwanted flip
     & \shortstack{Confident\\($p \geq 0.65$)}
     & \shortstack{Below threshold\\($p < 0.65$)} \\
\midrule
Valence & Unedited reconstruction
        & 133/214 (62.1\%) & 81/214 (37.9\%) & 14/81 (17.3\%) & 67/81 (82.7\%) \\
        & After physiological repair
        & 122/214 (57.0\%) & 92/214 (43.0\%) & 15/92 (16.3\%) & 77/92 (83.7\%) \\
\midrule
Arousal & Unedited reconstruction
        & 164/217 (75.6\%) & 53/217 (24.4\%) & 6/53 (11.3\%) & 47/53 (88.7\%) \\
        & After physiological repair
        & 170/217 (78.3\%) & 47/217 (21.7\%) & 7/47 (14.9\%) & 40/47 (85.1\%) \\
\bottomrule
\end{tabular}%
}
\end{table}

\subsection{Bandwave Continuity and Connectivity by Condition}
\label{app:bandwave}

Section~\ref{sec:continuity} summarizes boundary continuity. At each of the nine window boundaries, and for each electrode and band, we compute the amplitude jump, the absolute sample-to-sample step across the boundary divided by the median step over the eight neighbouring samples on either side (about 1 for continuous EEG), and the phase residual, the deviation of the Hilbert phase advance across the boundary from the advance expected from the neighbouring samples. Table~\ref{tab:seams} reports the medians over boundaries, electrodes and bands, pooled across both target directions.

\begin{table}[H]
\caption{Window-boundary continuity, median [Q1, Q3] over eligible recordings (valence 214, arousal 217). An amplitude jump near 1 indicates a continuous signal.}
\label{tab:seams}
\centering
\small
\setlength{\tabcolsep}{4pt}
\begin{tabular*}{\linewidth}{@{\extracolsep{\fill}}llccccc@{}}
\toprule
Task & Signal & Amplitude jump & $\theta$ & $\alpha$ & $\beta$ & Phase residual (rad) \\
\midrule
Valence & Recorded EEG & 1.04 [0.97, 1.10] & 1.03 & 1.05 & 1.02 & 0.84 \\
 & Unedited reconstruction & 1.74 [1.55, 1.99] & 2.00 & 1.80 & 1.34 & 1.25 \\
 & Target + latent proximity & 1.74 [1.55, 2.00] & 2.03 & 1.80 & 1.34 & 1.26 \\
 & Without physiology & 1.89 [1.68, 2.17] & 2.10 & 1.86 & 1.64 & 1.42 \\
 & Repair control & 2.09 [1.83, 2.34] & 2.36 & 2.14 & 1.63 & 1.39 \\
 & Base & 2.00 [1.77, 2.26] & 2.28 & 2.06 & 1.58 & 1.40 \\
 & Full typicality & 2.02 [1.78, 2.28] & 2.26 & 2.07 & 1.58 & 1.40 \\
\midrule
Arousal & Recorded EEG & 1.04 [0.97, 1.10] & 1.05 & 1.05 & 1.01 & 0.85 \\
 & Unedited reconstruction & 1.80 [1.56, 2.04] & 2.12 & 1.86 & 1.41 & 1.25 \\
 & Target + latent proximity & 1.80 [1.56, 2.04] & 2.11 & 1.88 & 1.41 & 1.25 \\
 & Without physiology & 2.07 [1.79, 2.35] & 2.30 & 2.03 & 1.70 & 1.42 \\
 & Repair control & 2.13 [1.89, 2.36] & 2.43 & 2.06 & 1.64 & 1.40 \\
 & Base & 2.07 [1.84, 2.32] & 2.38 & 2.04 & 1.59 & 1.40 \\
 & Full typicality & 2.07 [1.83, 2.33] & 2.39 & 2.06 & 1.65 & 1.40 \\
\bottomrule
\end{tabular*}
\end{table}

Across recordings, the reconstruction exceeds a jump of 1.5 in 78\% (valence) and 82\% (arousal) of cases, against none of the recorded trials, and full CFO adds a median paired increase of 0.24 over the reconstruction. The phase residual follows the same order: 0.84--0.85 rad for recorded EEG, 1.25 for the reconstruction and 1.40 for full CFO.

Table~\ref{tab:connectivity} reports coherence and debiased wPLI$^2$ with the componentwise estimator of Appendix~\ref{app:graded_physiology}, averaged over the 91 electrode pairs.
\begin{table}[H]
\caption{Connectivity by condition: median over eligible recordings of the mean coherence and debiased wPLI$^2$ across the 91 electrode pairs, pooled across both target directions.}
\label{tab:connectivity}
\centering
\small
\setlength{\tabcolsep}{4pt}
\begin{tabular}{@{}llcccccc@{}}
\toprule
& & \multicolumn{3}{c}{Coherence} & \multicolumn{3}{c}{Debiased wPLI$^2$} \\
\cmidrule(lr){3-5}\cmidrule(l){6-8}
Task & Signal & $\theta$ & $\alpha$ & $\beta$ & $\theta$ & $\alpha$ & $\beta$ \\
\midrule
Valence & Recorded EEG & 0.300 & 0.335 & 0.269 & 0.022 & 0.027 & 0.012 \\
 & Unedited reconstruction & 0.856 & 0.939 & 0.726 & 0.084 & 0.110 & 0.031 \\
 & Target + latent proximity & 0.856 & 0.939 & 0.726 & 0.084 & 0.110 & 0.031 \\
 & Without physiology & 0.856 & 0.939 & 0.726 & 0.084 & 0.110 & 0.031 \\
 & Repair control & 0.524 & 0.481 & 0.372 & 0.032 & 0.033 & 0.022 \\
 & Full typicality & 0.520 & 0.465 & 0.366 & 0.031 & 0.031 & 0.021 \\
\midrule
Arousal & Recorded EEG & 0.292 & 0.309 & 0.247 & 0.023 & 0.026 & 0.012 \\
 & Unedited reconstruction & 0.878 & 0.946 & 0.735 & 0.098 & 0.102 & 0.036 \\
 & Target + latent proximity & 0.878 & 0.946 & 0.735 & 0.098 & 0.102 & 0.036 \\
 & Without physiology & 0.878 & 0.946 & 0.735 & 0.097 & 0.102 & 0.036 \\
 & Repair control & 0.519 & 0.479 & 0.380 & 0.029 & 0.031 & 0.023 \\
 & Full typicality & 0.515 & 0.466 & 0.377 & 0.027 & 0.029 & 0.023 \\
\bottomrule
\end{tabular}
\end{table}

\subsection{Case Study in Relation to Other Participants}
\label{app:case_subjects}

We selected the case recordings from high-to-low full-CFO counterfactuals whose original prediction survived unedited reconstruction and whose decoded counterfactual met both target confidence and P10 acceptance (24 of 183 for arousal and 24 of 137 for valence). We ranked each quantity across all high-to-low recordings of the task and chose a recording with a small mean distance from the median rank over $d_z$, $\Delta_{\mathrm{dec}}$, $E_{\mathrm{rec}}$, the latent target probability, the decoded probability gain over $R(\mathbf Z)$, the physiology grade, the frontal alpha and overall beta amplitude changes, and the boundary jump. Figure~\ref{fig:subject_comparison} in Section~\ref{sec:case_study} shows the spread of these quantities by participant.

Over the 183 high-to-low arousal recordings, the median latent distance is $d_z=0.261$ [0.226, 0.299], $\Delta_{\mathrm{dec}}=0.294$ [0.275, 0.310], and the decoded probability gain over $R(\mathbf Z)$ is 0.147 [0.081, 0.264]. For the 137 valence recordings, the corresponding values are 0.266 [0.216, 0.309], 0.291 [0.271, 0.308] and 0.180 [0.108, 0.230]. The valence case recording (participant 20, trial 15) raised the low-valence probability from 0.495 to 0.925 in latent space and from 0.464 for $R(\mathbf Z)$ to 0.694 after decoding, at $d_z=0.246$ and $\Delta_{\mathrm{dec}}=0.287$. Its re-encoded discrepancy fell from 1.043 to 1.020 ($\tau_0=1.044$; the reconstruction was already accepted). It was VCSC-feasible with grade 0.992 and passed coherence and wPLI but not amplitude or band power. As for arousal, its amplitude change matched the repair endpoint (correlation 0.9997), which re-encoded at 0.501.

\begin{figure}[H]
    \centering
    \includegraphics[width=0.9\linewidth]{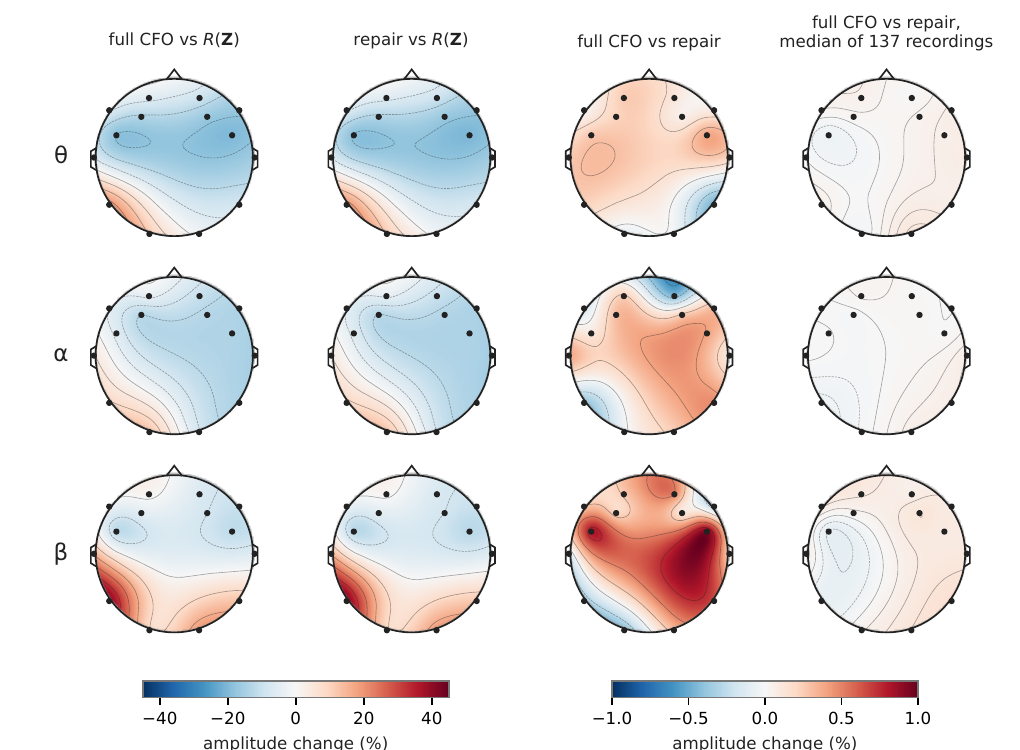}
    \caption{Band-amplitude change on the scalp for the valence case recording (participant 20, trial 15), as in Figure~\ref{fig:topomap_arousal}. The median in the last column is over the 137 high-to-low valence recordings.}
    \label{fig:topomap_valence}
\end{figure}

Figure~\ref{fig:topomap_valence} shows the same maps for the valence case recording. The full counterfactual and the repair endpoint again give nearly the same map (largest change 41\% in beta at P7), their difference stays below 0.9\%, and the median difference over the 137 high-to-low valence recordings is at most 0.15\%.

\paragraph{Edge masking.} For every eligible recording we split the decoded edit into the first and last 32 samples (0.25~s) of each window and the remaining 704 samples, added each part alone to its base signal, and re-encoded the result with the participant's frozen model. For full CFO the edit is the counterfactual minus the repair endpoint, added to the repair endpoint; for the condition without physiological penalties it is the counterfactual minus $R(\mathbf Z)$, added to $R(\mathbf Z)$. Re-encoding the unmasked signals reproduced the stored decoded probabilities to within $3\times10^{-4}$ and the Gaussian discrepancies to within $2\times10^{-5}$. Table~\ref{tab:edge_mask} covers recordings whose unedited reconstruction kept the original prediction and whose unmasked counterfactual received the target class. The valence case recording behaves like the arousal case: the edge-only edit keeps its probability of 0.694, and the middle-only edit returns it to the repair value of 0.501.

\begin{table}[H]
\caption{Edge masking for recordings whose unedited reconstruction kept the original prediction and whose unmasked counterfactual received the target class. Entries give target hits / target probability $\geq 0.65$ / P10 acceptance, with the median target probability in parentheses.}
\label{tab:edge_mask}
\centering
\small
\setlength{\tabcolsep}{4pt}
\begin{tabular}{@{}llrccc@{}}
\toprule
Task & Edit & $n$ & Whole edit & Edges only (0.25~s) & Middle only \\
\midrule
Valence & Full CFO $-$ repair & 109 & 109/43/73 (0.623) & 109/43/73 (0.623) & 29/1/5 (0.445) \\
 & Without physiology & 133 & 133/133/132 (0.993) & 133/133/132 (0.993) & 7/0/7 (0.424) \\
\midrule
Arousal & Full CFO $-$ repair & 93 & 93/35/45 (0.622) & 93/35/45 (0.622) & 10/0/0 (0.433) \\
 & Without physiology & 162 & 162/162/162 (0.995) & 162/162/162 (0.995) & 13/0/7 (0.383) \\
\bottomrule
\end{tabular}
\end{table}

To test the classifier directly, we took recorded trials from four participants per task (13 arousal trials from participants 3, 8, 12 and 19; 16 valence trials from participants 0, 5, 13 and 20) and replaced the samples between the outer $E$ samples of every window with those of another recorded trial of the same participant. With $E=64$ (0.5~s) the window probability changed by a median of $3\times10^{-4}$ (arousal) and $1\times10^{-4}$ (valence), and by at most 0.012; with $E=32$ by a median of 0.004 and 0.003; replacing whole windows changed it by a median of 0.15 and 0.18.

\subsection{Limitations}
\label{app:limitations}
The subject-adversarial objective may encourage the model to suppress features that identify individual participants, but it does not guarantee that all subject information is removed. Suppressing participant-identifying structure may also remove information useful for reconstruction; whether this explains the observed reconstruction errors remains a hypothesis. Moreover, the prediction model must be trained across LOSOCV folds for each affective dimension and counterfactual generation requires a separate iterative optimization for each evaluated trial. Therefore, reconstruction, classification, subject-adversarial training, and physiological constraints add significant computational cost. This limits the number of repeated runs, hyperparameter settings, ablations, and datasets that we can evaluate within this study. Self-report data introduce a significant challenge in modeling. Each six-second training window inherits its recording's rating. We evaluate emotion recognition and counterfactuals over the complete 60-second trial. No time-local affect labels are available, so a localized counterfactual edit cannot be interpreted as a validated momentary change in affect. DREAMER's valence and arousal labels are based on subjective five-point self-reports. Individual differences in interpreting the scale and the conversion of ratings into binary classes introduce uncertainty that the model cannot resolve. Although we may attribute specific EEG patterns to emotional classes, we are limited to the information and quality of the dataset.  While the methodology suggested here may be applied to other datasets, the performance and explanation quality may differ and require extra tuning.

\end{document}